\documentclass[letterpaper]{article} 
\usepackage{aaai2026}  
\usepackage{times}  
\usepackage{helvet}  
\usepackage{courier}  
\usepackage[hyphens]{url}  
\usepackage{graphicx} 
\usepackage{natbib}  
\usepackage{caption} 
\usepackage{algorithm}
\usepackage{algorithmic}

\usepackage{newfloat}
\usepackage{listings}
\DeclareCaptionStyle{ruled}{labelfont=normalfont,labelsep=colon,strut=off} 
\floatstyle{ruled}
\newfloat{listing}{tb}{lst}{}
\floatname{listing}{Listing}
\nocopyright 

\usepackage{amsmath,amsfonts,amsthm,bm,amssymb}
\usepackage{tikz}
\tikzset{every picture/.style={line width=0.75pt}} 

\usepackage{mathtools}
\usepackage{booktabs}
\usepackage{array}
\usepackage{multirow}
\usetikzlibrary{shadows} 

\newcommand{\indep}{\perp \!\!\! \perp}

\newtheorem{theorem}{Theorem}
\newtheorem{definition}[theorem]{Definition}
\newtheorem{proposition}[theorem]{Proposition}

\newtheorem{corollary}[theorem]{Corollary}
\newtheorem{lemma}[theorem]{Lemma}

\newtheorem{assumption}[theorem]{Assumption}

\definecolor{wongPurple}{RGB}{204, 121, 167}
\definecolor{wongLightBlue}{RGB}{86, 180, 233}
\definecolor{gris}{HTML}{A9A9A9}

\usetikzlibrary{shapes.geometric, arrows.meta, positioning, calc}

\usepackage[table]{xcolor} 
\usepackage{subcaption}
\usepackage{bm}
\usepackage{centernot}

\usepackage[textwidth=1.5cm, textsize=tiny]{todonotes}

\title{Decomposing Direct and Indirect Biases in Linear Models \\ under Demographic Parity Constraint}

\author{
    Bertille~Tierny\textsuperscript{\rm 1,2},
    Arthur~Charpentier\textsuperscript{\rm 3},
    François~Hu\textsuperscript{\rm 1}
}
\affiliations{
    \textsuperscript{\rm 1}Milliman France, R\&D Department, AI Lab\\
    \textsuperscript{\rm 2}ENSAE - Institut Polytechnique de Paris\\
    \textsuperscript{\rm 3}Université du Québec à Montréal\\
bertille.tierny@milliman.com, charpentier.arthur@uqam.ca, francois.hu@milliman.com
}

\begin{document}

\maketitle

\begin{abstract}
Linear models are widely used in high-stakes decision-making due to their simplicity and interpretability. Yet when fairness constraints such as demographic parity are introduced, their effects on model coefficients, and thus on how predictive bias is distributed across features, remain opaque. Existing approaches on linear models often rely on strong and unrealistic assumptions, or overlook the explicit role of the sensitive attribute, limiting their practical utility for fairness assessment.
We extend the work of \cite{chzhen2022minimax} and \cite{fukuchi2023demographic} by proposing a post-processing framework that can be applied on top of any linear model to decompose the resulting bias into direct (sensitive-attribute) and indirect (correlated-features) components. Our method analytically characterizes how demographic parity reshapes each model coefficient, including those of both sensitive and non-sensitive features. This enables a transparent, feature-level interpretation of fairness interventions and reveals how bias may persist or shift through correlated variables.
Our framework requires no retraining and provides actionable insights for model auditing and mitigation. Experiments on both synthetic and real-world datasets demonstrate that our method captures fairness dynamics missed by prior work, offering a practical and interpretable tool for responsible deployment of linear models.
\end{abstract}

\begin{links}
    \link{Code}{https://github.com/bias-mitigator/interpretable.git}
\end{links}


\section{Introduction}\label{sec:intro}
Linear models remain a foundational tool in statistical learning due to their interpretability, scalability, and simplicity \cite{hastie2009elements}. They are widely used in high-stakes domains such as credit scoring, hiring, insurance, and healthcare, where algorithmic decisions have significant consequences and fairness considerations are critical \citep{obermeyer2019dissecting, barocas2023fairness}. In these settings, linear models may inadvertently encode or amplify unfair biases. These biases can arise \textit{directly}, through the explicit use of sensitive attributes such as race or gender, or \textit{indirectly}, through features correlated with those attributes \cite{hajian2012methodology, nabi2018fair, tang2023and}. Fairness in machine learning has been extensively studied, with various formal definitions and mitigation strategies proposed \cite{del2020review, mehrabi2021survey, pessach2022review}. One of the most common criteria is \textit{Demographic Parity} (DP), which requires that the predictions be statistically independent of sensitive attributes. Although many methods aim to enforce DP in classification settings \cite{agarwal2018reductions, gaucher2023fair, hu2024sequentially}, few provide systematic tools to quantify and separate the sources of unfairness, especially in linear models. In particular, existing approaches, such as \cite{chzhen2022minimax,fukuchi2023demographic}, do not provide a systematic decomposition of bias stemming from the sensitive feature versus that induced by correlated non-sensitive features.
This lack of decomposition is especially problematic in linear models, which, despite their transparency, are not well understood in terms of how fairness constraints affect individual model coefficients. As a result, practitioners often lack insight into how these constraints redistribute predictive contributions across features or whether indirect biases persist even after sensitive variables are removed.

\subsection{Main Contributions}
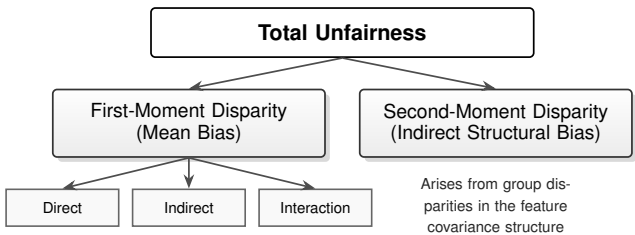
\begin{figure}[h]
\centering
\resizebox{\columnwidth}{!}{
\begin{tikzpicture}[
    node distance=0.5cm and 0.2cm, 
    font=\sf\small,
    main/.style={
        rectangle, draw=black, thick, rounded corners=2pt,
        text centered, text width=6cm, minimum height=0.8cm, font=\sf\bfseries
    },
    L1/.style={
        rectangle, draw=black!80, thick, rounded corners=2pt,
        top color=gray!1, bottom color=gray!15, 
        text centered, text width=4.2cm, minimum height=1.1cm, drop shadow={opacity=0.3}
    },
    L2/.style={
        rectangle, draw=black!60, fill=gray!5,
        text centered, text width=1.6cm, font=\sf\scriptsize, minimum height=0.6cm
    },
    detail/.style={
        font=\sf\scriptsize, 
        color=black!85,      
        text width=3.2cm,
        text centered
    },
    line/.style={-Stealth, thick, black!70}
]

\node (total) [main] {Total Unfairness};

\node (mean) [L1, below=of total, xshift=-2.5cm] {First-Moment Disparity \\(Mean Bias)};
\node (var)  [L1, below=of total, xshift=2.5cm] {Second-Moment Disparity (Indirect Structural Bias)};

\node (indirect) [L2, below=of mean]      {Indirect};
\node (direct)   [L2, left=of indirect]   {Direct};
\node (interact) [L2, right=of indirect]  {Interaction};

\node (var_detail) [detail, below=of var, yshift=0.3cm] {Arises from group disparities in the feature covariance structure};

\draw [line] (total.south) -- (mean.north);
\draw [line] (total.south) -- (var.north);

\draw [line] (mean.south) -- (direct.north);
\draw [line] (mean.south) -- (indirect.north);
\draw [line] (mean.south) -- (interact.north);

\end{tikzpicture}
} 
\caption{Conceptual decomposition of the total unfairness measure.The unfairness splits into two bias sources: disparities in the \textbf{mean} of predictions (First-Moment) and disparities in the \textbf{variance} of predictions (Second-Moment).}
\label{fig:unfairness_decomposition_sober}
\end{figure}

We propose a framework for learning fair linear models, designed to identify and mitigate both indirect and direct biases in linear models. Specifically:

\begin{itemize}
    \item We introduce a linear modeling framework aligned with standard practices and derive a closed-form solution for the optimal fair regressor. To our knowledge, this is the first solution that remains linear under group-wise feature standardization. In practice, it can be applied on top of any linear model (penalized, with or without intercept) making it broadly compatible and easily deployable.
    \item Building on this optimal solution, we disentangle the contributions of sensitive and non-sensitive features to fairness violations (see Fig.~\ref{fig:unfairness_decomposition_sober}) while providing clear guidance on how to adjust coefficients toward fairness.
    \item We illustrate the effectiveness of our approach on both synthetic and real-world datasets, demonstrating its ability to produce fair linear models while offering interpretability of both direct and indirect biases.
\end{itemize}

This work advances the understanding of fairness in linear models and contributes to the broader literature by providing tools to dissect and interpret bias at the feature level. For clarity of presentation, all proofs are provided in the supplementary materials.

\subsection{Related Work}
The study of fairness constraints in linear regression, particularly under DP, is relatively recent. Most existing methods either focus on model-level fairness objectives or rely on restrictive assumptions that limit their applicability in practice.

\cite{chzhen2022minimax} propose a minimax solution for linear regression under DP, deriving a closed-form intercept correction. However, their formulation is based on a strong assumption: the sensitive feature is independent of the other covariates. Therefore, they are omitting completely the indirect biases. This assumption rarely holds in real-world data and significantly restricts both the predictive accuracy of the model and the relevance of its fairness guarantees.

\cite{fukuchi2023demographic} extend this line of work by adjusting both intercept and non-sensitive feature coefficients. Although this allows more flexibility, their framework still omits an explicit treatment of the sensitive feature’s contribution, which limits bias diagnostics. Moreover, their solution also still builds on simplifying assumptions that may distort the fairness-performance trade-off.

In contrast, our approach explicitly characterizes the effect of DP constraints on all model components, including the sensitive feature. This enables a fine-grained decomposition of direct and indirect biases and provides clearer insights into how fairness interventions affect both predictive behavior and feature-level fairness contributions.
\begin{table}[H]
\centering
\resizebox{\columnwidth}{!}{%
\begin{tabular}{|l|c|c|c|c|}
\hline
& \textbf{Direct (Mean)} & \textbf{Indirect (Mean)} & \textbf{Interaction} & \textbf{Indirect (Structural)} \\
\hline
[CS22] & \checkmark &  & \checkmark & \\
\hline
[FS23] & \checkmark & \checkmark & \checkmark & \\
\hline
ours & \checkmark & \checkmark & \checkmark & \checkmark \\
\hline
\end{tabular}%
}
\caption{Comparison of bias mitigation methods across linear models proposed by [CS22] \cite{chzhen2022minimax}, [FS23] \cite{fukuchi2023demographic}, and our approach. Checkmarks indicate addressed biases.}
\end{table}

\subsection{Outline of the Paper} The remainder of this article is structured as follows: Section~\ref{sec:problem_formulation}, introduces the problem setup and the key metrics used throughout the article. Section~\ref{sec:sota_limitations} reviews the limitations of existing fair linear models. Section~\ref{sec:general_framework} presents our main contribution: a general framework for learning optimal fair linear models. This is followed in Section~\ref{sec:decompostion} by a decomposition of unfairness into direct and indirect biases. Finally, Section~\ref{sec:implementation} details the practical implementation of our methodology and Section~\ref{sec:numerical} presents numerical results comparing our method to state-of-the-art baselines.

\section{Problem Formulation}
\label{sec:problem_formulation}

Let $(\boldsymbol{X}, S, Y)$ be a random triplet, where $\boldsymbol{X} \in \mathcal{X} \subset \mathbb{R}^d$ is a non-sensitive feature vector, $Y \in \mathcal{Y} \subset \mathbb{R}$ is the target variable, and $S \in \mathcal{S}=[M]$ is a discrete sensitive attribute where $[M] := \{1, \dots, M\}$. We define $p_s=\mathbb{P}(S=s)$ for all $s\in [M]$. Additional notations are provided in Appendix~\ref{sub:Notations}. Our goal is to find a predictor $f: \mathcal{X} \times \mathcal{S} \to \mathcal{Y}$ from a set $\mathcal{F}$ that balances predictive utility with fairness. We denote by $\nu_f$ the distribution of $f(\boldsymbol{X}, S)$, and by $\nu_{f \mid s}$ its distribution given $S = s$. We make the following standard assumption.
\begin{assumption}
    For $f \in \mathcal{F}$, measures $(\nu_{f \mid s})_{s \in [M]}$ are non atomic with finite second moments.
\end{assumption}
We evaluate any predictor $f$ along three key and potentially competing dimensions: predictive risk, fairness, and goodness-of-fit. Each is formally defined below. 

\subsection{Measuring Risk}
We measure the predictive performance of a predictor using the classical quadratic risk, defined as:
\begin{equation*}
    \mathcal{R}(f) = \mathbb{E}\left[(f(\boldsymbol{X},S)-Y)^2\right].
\end{equation*}
This risk is uniquely minimized by the Bayes optimal predictor $f^*(\boldsymbol{X},S) = \mathbb{E}[Y\mid \boldsymbol{X},S]$, recognizing that fairness constraints entail a trade-off with this optimal benchmark.

\subsection{Measuring Unfairness}
Our work is grounded in the concept of Demographic Parity, which exists in both a weak and a strong form. In particular, a predictor $f$ satisfies \textit{Weak} DP if its expectation is independent of the sensitive attribute. That is,
$$
\mathbb{E}[f(\boldsymbol{X},S)\mid S=s] = \mathbb{E}[f(\boldsymbol{X},S)], \quad \text{for all } s \in [M],
$$
ensuring fairness at the level of the first moment (the mean).

\begin{definition}[(Strong) Demographic Parity]
\label{def:strong_dp}
A predictor $f$ satisfies Strong DP if its entire output distribution is independent of the sensitive attribute. That is,
$$
\nu_{f \mid s} = \nu_{f} \quad\text{for all }  s \in [M].
$$
\end{definition}
This is a much stricter criterion, requiring equivalence of all statistical moments.

\paragraph{Unfairness Measure}
We quantify unfairness through the lens of Strong DP, using Wasserstein-2 ($\mathcal{W}_2$) to measure distributional dissimilarities. For further details, we refer the reader to \cite{santambrogio2015optimal}. Specifically, the unfairness of a predictor $f$ is defined as the weighted sum of \( \mathcal{W}_2 \) distance between the group-conditional distributions \( (\nu_{f \mid s})_{s \in [M]} \) and their common barycenter:
\begin{flalign}
\label{eq:unfairness}
    \mathcal{U}(f)
    &=\min_{\nu \in \mathcal{P}_2(\mathbb{R})}\sum_{s=1}^M p_s \mathcal{W}_2^2(\nu_{f\mid s},\nu) \enspace.
\end{flalign}
A predictor $f$ is said to be exactly fair, that is, $\mathcal{U}(f) = 0$ \textit{iff} the predictor satisfies Strong DP. Thus, it provides a measure of how far a model is from achieving exact fairness.

\subsection{Measuring Goodness-of-fit}
Evaluating fair regression models requires more than assessing overall risk and unfairness. A key consideration is the group-conditional adequacy of the model. The classical coefficient of determination defined as $R^2(f)= \mathrm{Var}(f(\boldsymbol{X}, S)) / \mathrm{Var}(Y)$ is a standard metric for explained variance, particularly in linear settings. While it provides a familiar baseline, $R^2$ can obscure performance disparities and fails to capture group-specific \emph{goodness-of-fit}. For example, a linear model may approximate one group well but fit another poorly, a limitation not revealed by $R^2$.

\paragraph{Group-Weighted Coefficient of Determination ($GWR^2$).}
To diagnose this critical issue, we use the \emph{Group-Weighted $R^2$} ($GWR^2$). This metric is the average of the $R^2$ computed independently within each sensitive group, providing a direct measure of how well a model fits the data, on average, for all populations under consideration. For a predictor $f$, the definition is:
\begin{align*}
GWR^2(f) &:= \sum_{s \in \mathcal{S}} p_{s} R^2_s(f)\enspace,
\end{align*}
where,
$$
R^2_s = 1-\frac{\mathrm{Var}(Y - f(\boldsymbol{X}, s) \mid S = s)}{\mathrm{Var}(Y \mid S = s)}\enspace,
$$ 
The strength of this metric is theoretically grounded in our analysis of the gap between $GWR^2$ and the global $R^2$ (we refer to Appendix~\ref{app:gwr2_details} for further details). 
Divergence between these two metrics indicates model failure to capture group-specific structures. Thus, $GWR^2$ is a necessary diagnostic to signal structural mismatch that global metrics can obscure.

\section{Limitations of Existing Fair Linear Models}
\label{sec:sota_limitations}
The existing literature on fair linear regression provides foundational solutions but often relies on simplifying assumptions about the data-generating process. We review two key works that represent the progression from handling direct bias to incorporating some forms of indirect bias.

\paragraph{Mitigating Direct Bias.}
\cite{chzhen2022minimax} consider a hypothesis where unfairness arises solely from a group-dependent intercept term:
\begin{equation}\label{eq:Evgenii}
    Y = \langle \boldsymbol{X}, \boldsymbol{\beta}_{CS22} \rangle + \beta_{0,CS22}^{(s)}+\zeta, \quad \text{where } \zeta \sim \mathcal{N}(0,1),
\end{equation}
with the key assumption that features are independent of the sensitive group, i.e., $\boldsymbol{X} \indep S$. In this setting, the associated Bayes optimal predictor is $\langle \boldsymbol{X}, \boldsymbol{\beta}_{CS22}\rangle + \beta_{0,CS22}^{(s)}$. The independence assumption eliminates all sources of indirect bias by construction, isolating direct bias as the only source of unfairness. Therefore, achieving fairness is straightforward.

\begin{lemma}[Adapted from \cite{chzhen2022minimax}]
    Given the equation in Eq.~\eqref{eq:Evgenii}, the optimal DP-fair predictor is obtained by averaging out the group-specific intercepts:
$$
f_{CS22}(\boldsymbol{x},s)=\langle\boldsymbol{x},\boldsymbol{\beta}_{CS22}\rangle + \sum_{s\in[M]}p_s\beta_{0,CS22}^{(s)}.
$$
\end{lemma}

\paragraph{Mitigating Indirect Mean Bias.}
\cite{fukuchi2023demographic} relax the feature independence assumption, allowing for group-dependent feature means and slopes:
\begin{equation}\label{eq:Riken}
    Y = \langle \boldsymbol{X}, \boldsymbol{\beta}^{(S)}_{FS23} \rangle +\zeta,  \quad \text{where } \zeta \sim \mathcal{N}(0,1),
\end{equation}
where $\boldsymbol{X} \sim \mathcal{N}(\boldsymbol{\mu}^{(s)}, \sigma_X^2I)$. This structure introduces an indirect bias that results from the differing feature means $\boldsymbol{\mu}^{(s)}$. However, it maintains a restrictive assumption of homoscedastic, uncorrelated features across groups.

\begin{lemma}[Adapted from \cite{fukuchi2023demographic}, Lemma 1]
Given the model in Eq.~\eqref{eq:Riken}, the optimal DP-fair predictor is: 
\begin{multline*}
    {f_{FS23}}(\boldsymbol{x},s)=\| {\boldsymbol{\beta}^{(.)}_{FS23}}\| \langle \tilde{\boldsymbol{\beta}}^{(s)}_{FS23},\boldsymbol{x} -\boldsymbol{\mu}^{(s)}\rangle \\+ \sum_{s' \in [M]}{p}_{s'} \langle \boldsymbol{\beta}^{(s')}_{FS23},\boldsymbol{\mu}^{(s')}\rangle ,
\end{multline*}
with
\[
\|\boldsymbol{\beta}^{(.)}_{FS23}\|=\sum_{s \in [M]}p_{s}\|\boldsymbol{\beta}^{(s)}_{FS23}\|\quad \text{and} \quad \tilde{\boldsymbol{\beta}}^{(s)}_{FS23}=\frac{\boldsymbol{\beta}^{(s)}_{FS23}}{\| \boldsymbol{\beta}^{(s)}_{FS23}\|}.
\]
\end{lemma}

\paragraph{Limitations of Prior Work} While these works represent important progress, they rely on restrictive assumptions about the data covariance structure. In particular, they do not address heteroscedasticity, where the feature covariance matrix $\Sigma^{(s)}$ varies across groups. As a result, it overlooks \textit{indirect structural bias} from distributional disparities, highlighting the need for a more general approach.

\section{A General Framework for Optimal Fair Regression}
\label{sec:general_framework}
We introduce a linear model framework that captures all key sources of bias, enabling us to derive the optimal fair predictor for more complex, group-dependent data structures.

\subsection{The General Model}
We consider a setting where the outcome $Y$ is generated by:
\begin{equation}\label{eq:our_model}
    Y = \langle \boldsymbol{X}, \boldsymbol{\beta^*} \rangle + \gamma^* S + \beta_0^* + \zeta ,
\end{equation}
where the features $\boldsymbol{X} \mid S=s \sim \mathcal{N}(\boldsymbol{\mu}^{(s)}, \mathbf{\Sigma}^{(s)})$ are group-dependent, and the noise $\zeta \sim \mathcal{N}(0,1)$ is independent of $S$ and $\boldsymbol{X}$. This model captures direct bias ($\gamma^*$), indirect mean bias ($\boldsymbol{\mu}^{(s)}$), and indirect structural bias ($\boldsymbol{\Sigma}^{(s)}$).

Our goal is to find the optimal predictor within the class of linear models, $\mathcal{F}_{\text{linear}}$, that minimizes the quadratic risk $\mathcal{R}$ subject to Strong DP. Given $(\boldsymbol{x},s) \in \mathcal{X}\times \mathcal{S}$, the Bayes optimal predictor is $f^*(\boldsymbol{x},s) = \langle \boldsymbol{x}, \boldsymbol{\beta^*} \rangle + \gamma^* s + \beta_0^*$.

\subsection{The Optimal Risk-Fairness Trade-off}
We seek to find the predictor that optimally navigates the trade-off between minimizing risk and ensuring fairness.
To formalize this, we adopt the $\varepsilon$-\textit{Relative Fairness Improvement} ($\varepsilon$-RI) constraint from \cite{chzhen2022minimax}. A predictor $f_\varepsilon$ satisfies this constraint if its unfairness is bounded by an $\varepsilon$-fraction of the Bayes-optimal predictor: 
$$
\mathcal{U}(f_\varepsilon) \le \varepsilon \cdot \mathcal{U}(f^*)\enspace.
$$
A key result, applicable to our framework, is that the predictor achieving the optimal risk-fairness trade-off under this constraint, \textit{i.e.}, verifying $f^*_\varepsilon \in \arg \min \{\mathcal{R}(f):\mathcal{U}(f)\leq \varepsilon \cdot \mathcal{U}(f^*)\}$, is a linear interpolation of the Bayes predictor $f^*$ and the optimal fair predictor $f^*_{DP}$:
\begin{equation*}
\label{eq:ri_interpolation}
    f^*_\varepsilon = (1-\sqrt{\varepsilon})f^*_{DP} + \sqrt{\varepsilon}f^* \enspace.
\end{equation*}
Our main result is to derive the explicit closed-form expression for $f^*_\varepsilon$ within our Gaussian linear model framework.

\subsection{Characterizing the Optimal Fair Predictor}

To state our main result, we first define the group-conditional mean and standard deviation of the Bayes optimal score:
\begin{itemize}
    \item Group-conditional mean: 
    $$\mu_{f^*}^{(s)} := \mathbb{E}[f^*(\boldsymbol{X},S)\mid S=s] = \langle \boldsymbol{\mu}^{(s)}, \boldsymbol{\beta^*} \rangle + \gamma^*s + \beta_0^*.$$
    \item Group-conditional variance: $$(\sigma_{f^*}^{(s)})^2 := \mathrm{Var}(f^*(\boldsymbol{X},S)\mid S=s) = (\boldsymbol{\beta^*})^\top\boldsymbol{\Sigma^{(s)}}\boldsymbol{\beta^*}.$$
\end{itemize}
We also define their population-level averages, weighted by the group prior probabilities $p_s$:
$$
\bar{\mu}_{f^*} = \sum_{s' \in [M]}p_{s'}\mu_{f^*}^{(s')}\quad \text{and} \quad
\bar{\sigma}_{f^*} = \sum_{s' \in [M]}p_{s'}\sigma_{f^*}^{(s')}
\enspace.
$$

\begin{proposition}[Optimal $\varepsilon$-Fair Predictor]
\label{prop:optimal_fair_predictor}
For the model in Eq.~\eqref{eq:our_model}, the unique predictor $f^*_\varepsilon$ that satisfies the $\varepsilon$-RI constraint and minimizes the quadratic risk is given by:
\begin{equation}
    f^*_\varepsilon(\boldsymbol{x},s) = \sigma_{\varepsilon}^{(s)} \left( \frac{\langle \boldsymbol{x} - \boldsymbol{\mu}^{(s)}, \boldsymbol{\beta^*} \rangle}{\sigma_{f^*}^{(s)}} \right) + \mu_{\varepsilon}^{(s)} \enspace,
\end{equation}
where the mean and std are convex combinations of the group-specific and population-averaged statistics:
\begin{align*}
    \mu_{\varepsilon}^{(s)} &= (1-\sqrt{\varepsilon}) \bar{\mu}_{f^*} + \sqrt{\varepsilon}\mu_{f^*}^{(s)} \\
    \sigma_{\varepsilon}^{(s)} &= (1-\sqrt{\varepsilon}) \bar{\sigma}_{f^*} + \sqrt{\varepsilon}\sigma_{f^*}^{(s)} \enspace.
\end{align*}
The optimal exactly-fair predictor $f^*_{DP}$ is recovered at $\varepsilon=0$, and the Bayes optimal predictor $f^*$ is recovered at $\varepsilon=1$.
\end{proposition}

\subsection{Interpreting the Fairness Mechanism}
The structure of $f^*_\varepsilon$ reveals a clear and tunable mechanism for enforcing fairness, which can be understood from two complementary perspectives.

\paragraph{Perspective 1: Tunable Standardization and Averaging.}
This perspective views fairness as the controlled shift of group-dependent moments toward global average moments.
\begin{enumerate}
    \item \textbf{Group-wise Standardization:} within each group $s$, the term $\langle \boldsymbol{x} - \boldsymbol{\mu}^{(s)}, \boldsymbol{\beta^*} \rangle / \sigma_{f^*}^{(s)}$ creates a standardized score (zero mean and unit variance). This procedure simultaneously removes indirect mean and structural biases.
    \item \textbf{Controlled Re-scaling and Shifting:} This standardized score is then re-scaled by $\sigma_{\varepsilon}^{(s)}$ and shifted by $\mu_{\varepsilon}^{(s)}$. 
    These coefficients are a direct interpolation between the group-specific moments ($\mu_{f^*}^{(s)}, \sigma_{f^*}^{(s)}$) and the global averages ($\bar{\mu}_{f^*}, \bar{\sigma}_{f^*}$). The parameter $\varepsilon$ directly control this trade-off: at $\varepsilon=0$, the predictor uses only global averages, eliminating all bias; at $\varepsilon=1$, it uses only group-specific values, retaining all original bias for maximum accuracy.
\end{enumerate}

\paragraph{Perspective 2: A Group-Conditional Fair Model.}
Alternatively, we can express the predictor as a linear model, 
$$
f^*_\varepsilon(\boldsymbol{x},s) = \langle \boldsymbol{x}, \boldsymbol{\beta}_{\varepsilon}^{(s)} \rangle + \beta_{0,\varepsilon}^{(s)}\enspace,
$$
to see how fairness is encoded into the parameters of the model.
By rearranging the terms from Proposition~\ref{prop:optimal_fair_predictor}, we find the effective slope and intercept for each group are:
\begin{align*}
    \boldsymbol{\beta}_{\varepsilon}^{(s)} = \left(\frac{\sigma_{\varepsilon}^{(s)}}{\sigma_{f^*}^{(s)}}\right) \boldsymbol{\beta}^* \quad \text{and} \quad
    \beta_{0,\varepsilon}^{(s)} = \mu_{\varepsilon}^{(s)} - \left(\frac{\sigma_{\varepsilon}^{(s)}}{\sigma_{f^*}^{(s)}}\right) \langle \boldsymbol{\mu}^{(s)}, \boldsymbol{\beta^*} \rangle \enspace.
\end{align*}
This view highlights that fairness is achieved by constructing a group-aware model with parameters systematically adjusted to counteract group-specific biases. The scaling factor $\sigma_{\varepsilon}^{(s)} / \sigma_{f^*}^{(s)}$ compensates for the structural bias, while the intercept $\beta_{0,\varepsilon}^{(s)}$ corrects for the mean-based biases.

\section{Decomposition of direct and indirect biases through the unfairness}
\label{sec:decompostion}

In this section, we develop a comprehensive framework for understanding unfairness in linear regression.  

\subsection{Prediction-level Decomposition of Unfairness}

We begin by decomposing our unfairness measure $\mathcal{U}(f)$ for any predictor within the class of linear models, $\mathcal{F}_{\text{linear}}$. 

\begin{proposition}[Linear Model Bias Decomposition]
\label{prop:bias_decomp}
For any predictor $f \in \mathcal{F}_{\text{linear}}$ with coefficients $(\boldsymbol{\beta}, \gamma, \beta_0)$, its total unfairness $\mathcal{U}(f)$ decomposes into First-Moment Disparity (FMD) and Second-Moment Disparity (SMD):
\begin{equation}
\mathcal{U}(f) = \underbrace{\mathrm{Var}(\mathbb{E}[f|S])}_{\text{FMD}} + \underbrace{\mathrm{Var}(\sqrt{\mathrm{Var}(f|S)})}_{\text{SMD}}.
\end{equation}
These components further decompose into four bias sources:
\begin{align}
\mathcal{U}(f) = \underbrace{\gamma^2 \mathrm{Var}(S)}_{\text{Direct Mean}}
    &+ \underbrace{\mathrm{Var}(\langle \boldsymbol{\mu}^{(S)}, \boldsymbol{\beta} \rangle)}_{\text{Indirect Mean}}
    + \underbrace{2\gamma \mathrm{Cov}(S, \langle \boldsymbol{\mu}^{(S)}, \boldsymbol{\beta} \rangle)}_{\text{Interaction}} \nonumber \\ 
    &+ \underbrace{\mathrm{Var}\left(\sqrt{\boldsymbol{\beta}^\top\boldsymbol{\Sigma}^{(s)}\boldsymbol{\beta}}\right)}_{\text{Indirect Structural}}.
\end{align}
\end{proposition} 

This decomposition formalizes the conditions required to achieve Strong DP, showing that fairness in this stronger sense necessitates mitigating bias at two distinct levels:
\begin{itemize}
    \item The \textbf{First-Moment Disparity} $\mathrm{Var}(\mathbb{E}[f\mid S])$ captures unfairness in average predictions. It arises from direct dependence on the sensitive attribute (Direct Mean Bias, related to Weak DP) or from correlations between group membership and feature means (Indirect Mean Bias).

    \item The \textbf{Second-Moment Disparity} $\mathrm{Var}(\sqrt{\mathrm{Var}(f\mid S)})$ captures a more subtle form of unfairness (Indirect Structural Bias) where predictive certainty differs across groups due to variations in feature covariance $\boldsymbol{\Sigma}^{(s)}$.
\end{itemize}
This decomposition reveals that a model can satisfy Weak DP (without FMD) while remaining unfair under Strong DP. The following corollary demonstrates a key advantage of our optimal $\varepsilon$-fair predictor:
\begin{corollary}[Residual Unfairness of our method]
\label{cor:unfairness_epsilon}
The total unfairness of our predictor $f^*_\varepsilon$, (see Prop.~\ref{prop:optimal_fair_predictor}), is exactly:
\begin{multline*}
    \mathcal{U}(f^*_\varepsilon) = \varepsilon \cdot \mathrm{Var}(\mathbb{E}[f^*\mid S])+ \varepsilon \cdot \mathrm{Var}(\sqrt{\mathrm{Var}(f^*\mid S)})\enspace.
\end{multline*}
\end{corollary}
This corollary highlights a direct, analytical link between a single control parameter ($\varepsilon$) and the total amount of multi-source unfairness, a property not available in prior models.

\subsection{Feature-level Decomposition of Unfairness via Approximation}

While the prediction-level decomposition quantifies total unfairness, practical intervention requires attributing this unfairness to individual features. A fully additive decomposition is challenging due to the nonlinearity introduced by the square root in the structural bias. To enable interpretability, we apply a first-order Taylor expansion to linearize this term, yielding a tractable and accurate additive approximation.

\paragraph{The Additive Case: Uncorrelated Features.} We consider a simplified setting where features are mutually uncorrelated within each group ($\boldsymbol{\Sigma}^{(s)}$ are diagonal matrices). In this case, the total indirect unfairness of any linear model decomposes into a sum of marginal contributions from each feature.

\begin{proposition}[Additive Feature-Level Decomposition]
\label{prop:feature_decomp_diag}
Given $f \in \mathcal{F}_{\text{linear}}$ with coefficients $(\boldsymbol{\beta}, \gamma)$, let its indirect unfairness be $\mathcal{U}_{\mathrm{indirect}}(f) = \mathcal{U}(f) - \gamma^2\mathrm{Var}(S)$. If all $\Sigma^{(s)}$ are diagonal, then this unfairness can be approximated by an additive sum:
$$
\mathcal{U}_{\mathrm{indirect}}(f) \approx \sum_{j=1}^{d} \mathcal{U}^{\text{approx}}_j(f),
$$
with the approximate main contribution from feature $X_j$ is:
\begin{align*} \label{eq:marginal_contrib_diag_approx}
\mathcal{U}^{\text{approx}}_j(f) = \underbrace{(\beta_j)^2 \mathrm{Var}(\mu_j^{(S)})}_{\text{Mean}}
            &+ \frac{1}{4\bar{V}} \underbrace{(\beta_j)^4 \mathrm{Var}((\sigma_j^{(S)})^2)}_{\text{Structural}}
            \\ +  &\underbrace{2\gamma \beta_j \mathrm{Cov}(S, \mu_j^{(S)})}_{\text{Interaction}}, \nonumber
\end{align*}
where $\mu_j^{(s)} = \mathbb{E}[X_j|S=s]$ and $(\sigma_j^{(s)})^2 = \mathrm{Var}(X_j|S=s)$. Here, $\bar{V} = \mathbb{E}[\mathrm{Var}(f|S)]$ is the average conditional score variance. 
\end{proposition}
This proposition attributes model unfairness to individual features via three pathways: (1) mean disparity, (2) variance disparity (structural bias), and (3) interaction with direct bias. The term $1/(4\bar{V})$ indicates that structural bias diminishes as predictive variance increases.

\paragraph{The General Case: Interactional Unfairness.}
When features are correlated, the decomposition becomes more complex due to cross-terms capturing \textit{interactional unfairness} (see Appendix). This includes: (1) the compounding of mean biases through correlated feature means, and (2) a deeper structural effect, which we term \textit{Covariance Disparity}, driven by group-level differences in feature correlations.

This analysis provides both practical and comprehensive insight. The additive decomposition highlights features with primary unfairness, while the general case reveals how feature correlations amplify or mitigate these effects.
\begin{figure*}[h!]
    \centering
    \includegraphics[width=0.9\linewidth]
    {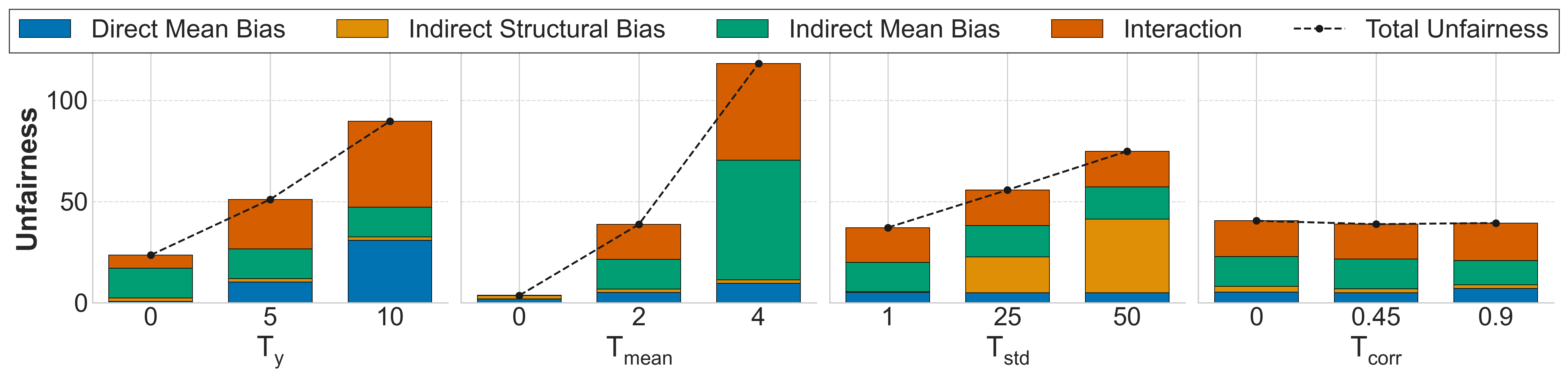}
    \caption{Bias decomposition (see Prop.~\ref{prop:bias_decomp}) of the linear model on synthetic data using by default T $= (3, 2, 3, 0.7)$.}
    \label{fig: Linear model's biases decomposition.}
\end{figure*}
\begin{figure*}[htbp!]
    \centering
    \includegraphics[width=0.95\linewidth]
    {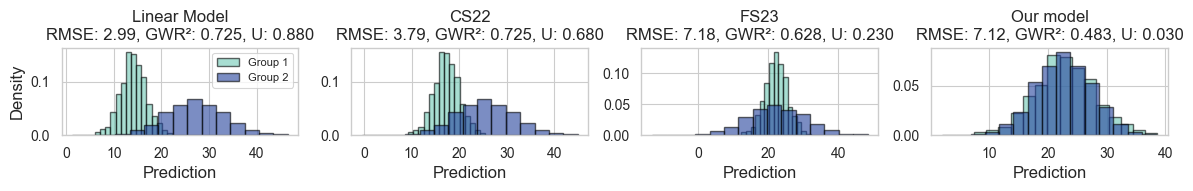}
    \caption{Comparison of group-conditioned model output distribution on synthetic data using T $= (10, 2, 2, 0.7)$.}
    \label{fig: Scores distributions on simulated data.}
\end{figure*}

\section{Practical Implementation and Estimation}
\label{sec:implementation}

To apply our framework in practice, the optimal fair predictor must be estimated from finite data, since the population parameters $(\boldsymbol{\beta}^*, \gamma^*, \boldsymbol{\mu}^{(s)}, \boldsymbol{\Sigma}^{(s)})$ are unknown. 

\paragraph{The Plug-in Estimator}

The plug-in estimator $\hat{f}_\varepsilon$ of $f^*_\varepsilon$ is constructed by replacing all quantities in Prop.~\ref{prop:optimal_fair_predictor} with their empirical estimates. 
\begin{enumerate}
    \item \textbf{Estimate Model Parameters}. We estimate the base model parameters $(\hat{\boldsymbol{\beta}}, \hat{\gamma}, \hat{\beta}_0)$. Our framework is agnostic to the fitting procedure; any standard method, such as OLS or penalized version (Ridge, Lasso), is applicable.

    \item \textbf{Estimating Group Statistics}. For each $s$, we compute the standard estimates for the group proportions $\hat{p}_s$, feature means $\hat{\boldsymbol{\mu}}^{(s)}$, and feature covariance matrices $\hat{\boldsymbol{\Sigma}}^{(s)}$.
    \item \textbf{Assemble the Fair Predictor.} Finally, these empirical components are used to construct the plug-in versions of the conditional score moments ($\hat{\mu}_{f}^{(s)}$, $\hat{\sigma}_{f}^{(s)}$) and their population averages ($\hat{\bar{\mu}}_{f}$, $\hat{\bar{\sigma}}_{f}$). These are then combined according to Prop.~\ref{prop:optimal_fair_predictor} to form the final estimator. 

\end{enumerate}

\paragraph{Evaluation Metrics}

We evaluate all models on a held-out test set using empirical estimators of our three key metrics.
For both the Risk and \( GWR^2 \), we consider their empirical counterparts, denoted \( \hat{\mathcal{R}} \) (mean squared error) and \( \widehat{GWR^2} \), respectively, where:
$$
\widehat{GWR^2}(f) = \sum_{s \in [M]} \hat{p}_s \left( 1 - \frac{\widehat{\mathrm{Var}}(Y - f \mid S=s)}{\widehat{\mathrm{Var}}(Y \mid S=s)} \right).
$$
We quantify the unfairness using the Kolmogorov-Smirnov (KS) test, as it is model-agnostic and does not rely on structural assumptions.
    \[ \hat{\mathcal{U}}_{\text{KS}}(f) = \max_{s_j, s_k \in [M]} D_{\text{KS}}(\hat{F}_{f\mid s_j}, \hat{F}_{f\mid s_k}). \]
    Here, $\hat{F}_{f|s}$ is the empirical CDF of scores for group $s$.

\section{Numerical Experiments} 
\label{sec:numerical}

We run experiments on synthetic and real-world data to: (1) validate our bias decomposition framework and (2) demonstrate that our tunable predictor, $\hat{f}_\varepsilon$, effectively traces the optimal risk-fairness frontier, outperforming prior methods.

\begin{figure}[htbp!]
    \centering
    \includegraphics[width=0.95\linewidth]{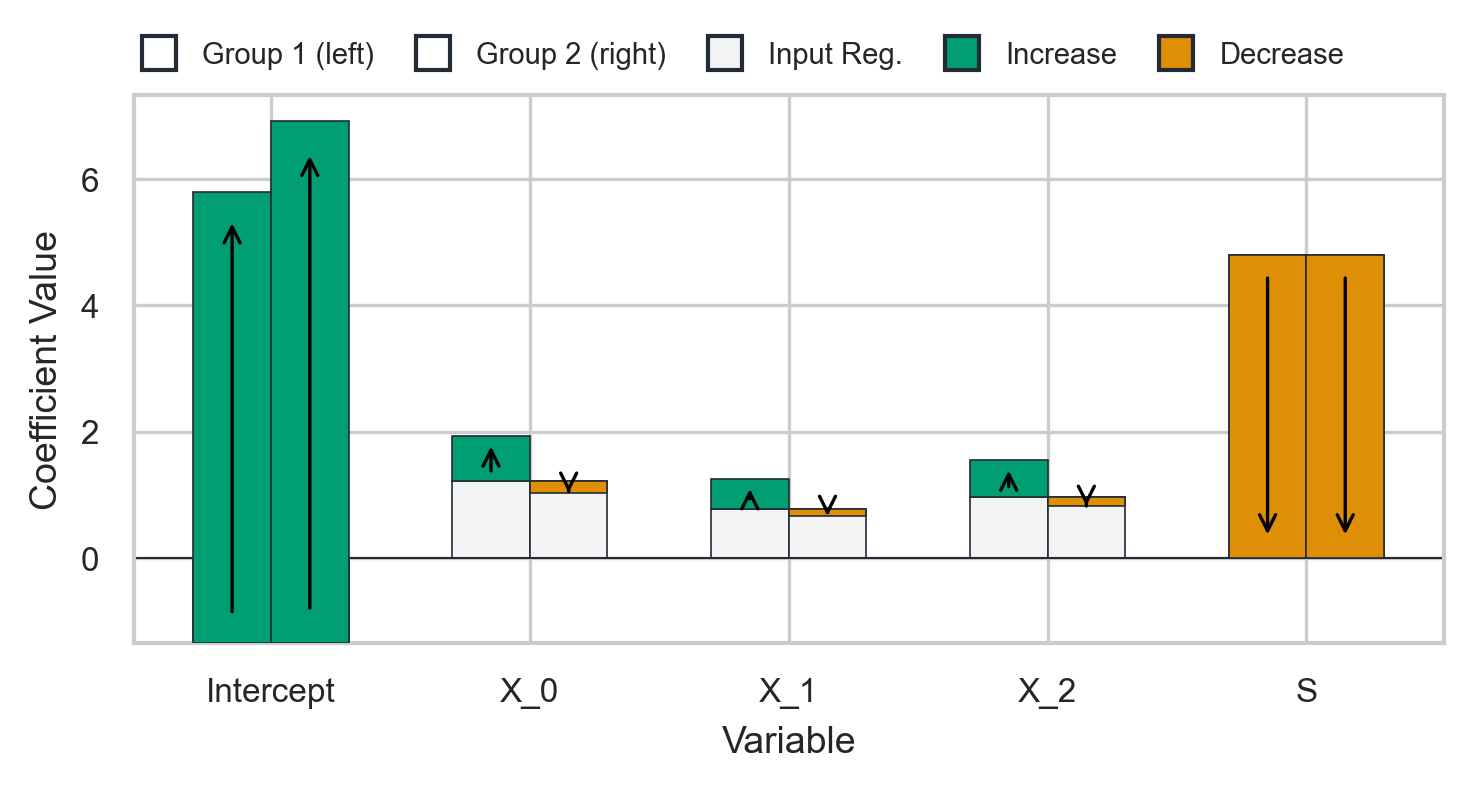}
    \caption{Coefficients adjustments for fairness, shown for a sample of features on synthetic data with T $=(3,2,3,.7)$.}
    \label{fig: Coefficients adjustments.}
\end{figure}

\subsection{Application on Synthetic Data} \label{sub: synthetic data description}

We generated synthetic
triplets $(\boldsymbol{X}, S, Y )$ where we can precisely control each source of bias. The sensitive attribute $S \in \{1,2\}$ is drawn from a Bernoulli distribution. The data-generating process is governed by four control parameters $T :=(T_y,T_{\text{mean}},T_{\text{std}}, T_{\text{corr}})$ that map directly to our bias decomposition: $T_y$ sets the \textbf{direct bias} coefficient $\gamma^*$; $T_{\text{mean}}$ introduces \textbf{indirect mean bias} by creating differences between feature means $\boldsymbol{\mu}^{(s)}$; and $T_{\text{std}}$ and $T_{\text{corr}}$ introduce \textbf{indirect structural bias} by creating group-specific differences in the variances and correlations within the covariance matrix $\boldsymbol{\Sigma}^{(s)}$. When a parameter is set to zero, the corresponding source of bias is absent. We provide full implementation details on the simulation of ($\boldsymbol{X}$, S, Y) in the Appendix~\ref{sub: synthetic data}.

\begin{table*}[h]
\centering
{\fontsize{9}{9}\selectfont 
\setlength{\tabcolsep}{1mm} 
\begin{tabular}{l|ccc|ccc|ccc}
\hline
\multirow{2}{*}{\textbf{Model}} & \multicolumn{3}{c|}{\textbf{CRIME}} & \multicolumn{3}{c|}{\textbf{LAW}} & \multicolumn{3}{c}{\textbf{GOSSIS}} \\
\cline{2-10}
 & \textbf{GWR$^2$} & \textbf{RMSE} & \textbf{Unfairness} & \textbf{GWR$^2$} & \textbf{RMSE} & \textbf{Unfairness} & \textbf{GWR$^2$} & \textbf{RMSE} & \textbf{Unfairness} \\
\hline
Base Model Unaware    & .45 $\pm$ .05 & 0.15 $\pm$ 0.01 & 0.55 $\pm$ 0.04
                & .15 $\pm$ .01 & 0.37 $\pm$ .00 & .13 $\pm$ .01
                & .69 $\pm$ .01 & 10.3 $\pm$ 0.1 & .14 $\pm$ .01
                \\
Base Model    & .46 $\pm$ .05 & 0.15 $\pm$ 0.01 & 0.61 $\pm$ 0.04
                & .15 $\pm$ .01 & 0.37 $\pm$ .00 & .43 $\pm$ .02
                & .69 $\pm$ .01 & 10.3 $\pm$ 0.1 & .15 $\pm$ .01
                \\
\textsc{CS22} & .46 $\pm$ .05 & 0.15 $\pm$ 0.01 & 0.54 $\pm$ 0.04
        & .15 $\pm$ .01 & 0.37 $\pm$ .00 & .08 $\pm$ .01
        & .69 $\pm$ .01 & 10.3 $\pm$ 0.1 & .14 $\pm$ .01
        \\
\textsc{FS23} & .35 $\pm$ .09 & 0.19 $\pm$ 0.01 & 0.20 $\pm$ 0.05
        & .08 $\pm$ .05 & 0.39 $\pm$ .01 & .15 $\pm$ .05
        & .51 $\pm$ .40 & 12.5 $\pm$ 3.3 & .13 $\pm$ .07
        \\
\textbf{Our model}& .38 $\pm$ .07 & 0.19 $\pm$ 0.01 & \cellcolor{green!15} \textbf{0.12 $\pm$ 0.04}
          &.15 $\pm$ .01 & 0.37 $\pm$ .00 & \cellcolor{green!15} \textbf{.07 $\pm$ .02}
          & .69 $\pm$ .01 & 10.4 $\pm$ 0.1 & \cellcolor{green!15} \textbf{.03 $\pm$ .01}
          \\
\hline
\end{tabular}}
\caption{Comparison of model performances across all datasets. Results are presented as mean $\pm$ standard deviation over 50 runs. 
Bold cells indicate the lowest unfairness.}
\label{tab:model_comparison}
\end{table*}

\paragraph{Experimentation scheme} 
Default parameters are set as follows : $d=5$, $\tau = 0.6$. Given the vector $T=(T_y$, $T_{\text{mean}}$, $T_{\text{std}}$,$T_{\text{corr}})$, we create synthetic dataset of $n=20,000$ samples and split it into training (50\%), testing (25\%), and unlabeled (25\%) subsets. As a base model, we chose the linear regression of $Y$ on $\boldsymbol{X}$ and $S$, using standard parameters from \texttt{scikit-learn} in Python. The coefficients of this linear regression are used as an input to build of fair linear model.

\begin{figure}[htbp!]
    \centering
    \includegraphics[width=0.9\columnwidth]{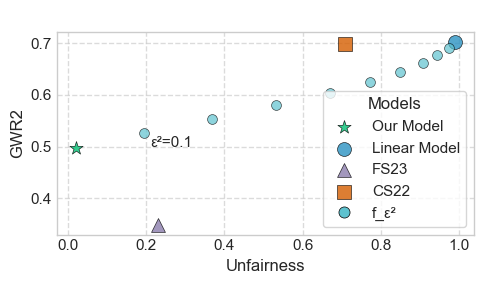}
    \caption{Analysis of Approximate fairness model on synthetic data with T $=(10, 2, 3, 0.7)$.}
    \label{fig:approximate:fairness}
\end{figure}

\paragraph{Validating the Bias Decomposition} Our framework provides a direct way to diagnose the sources of unfairness. Fig.~\ref{fig: Linear model's biases decomposition.} applies our decomposition to a linear model trained on synthetic data. The results empirically validate our theory: increasing the direct bias parameter ($T_y$) primarily inflates the Direct Mean and Interaction terms, while increasing the indirect parameters ($T_{\text{mean}}, T_{\text{std}}$) maps clearly to the Indirect Mean and Indirect Structural bias components, respectively. This confirms that our decomposition is a practical tool for identifying the root causes of unfairness in linear models.

\paragraph{Fairness Mitigation and Robustness to Bias Shifts.} 
In complex scenarios with full bias interactions $(3,2,2,0.7)$, our model uniquely preserves remediation capabilities (Fig.~\ref{fig: Scores distributions on simulated data.}). 
The remediation operates through three visible mechanisms (Fig.~\ref{fig: Coefficients adjustments.}): (1) direct bias elimination via sensitive attribute coefficient nullification, (2) structural bias remediation through feature coefficient adjustments, and (3) intercept compensation for all bias corrections. We refer to Appendix~\ref{sub: appendix numerical experiments} for additional analyses. 

\begin{figure}[htbp!]
    \centering
    \includegraphics[width=0.9\columnwidth]{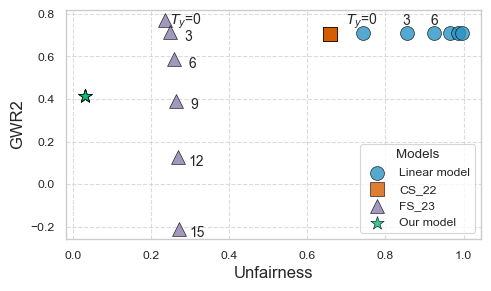}
        \caption{Analysis of Model performance \textit{w.r.t.} direct bias shifts ($T_y$) on synthetic data using T $=(*,2,2,0.7)$.}
        \label{fig:T_y varie, autres bias actifs}
\end{figure}

We also test robustness under bias shifts by varying \( T_y, T_{\text{mean}}, T_{\text{corr}}, T_{\text{std}} \). As shown in Fig.~\ref{fig:T_y varie, autres bias actifs}, our method and \textsc{CS22} remain stable in both performance and fairness as direct bias (\( T_y \)) increases, while \textsc{FS23} deteriorates.

\paragraph{Tracing the Optimal Risk–Fairness Frontier.}
Under the \( \varepsilon \)-RI constraint, the parameter \( \varepsilon \) provides continuous control over the desired fairness level. Fig.~\ref{fig:approximate:fairness} shows that our method consistently dominates in this bias scenario: it either achieves higher accuracy than the baselines at a given unfairness level, or ensures lower unfairness at a fixed accuracy, effectively tracing the optimal risk-fairness trade-off.

\subsection{Results on Real-World Data}

We use three standard and diverse fairness benchmarks.

GOSSIS \cite{raffa2022global}: contains medical information from over 130,000 patients admitted to intensive care units. The task consists in predicting the vital variable \textit{h1\_diaspb\_max} using ethnicity as a protected attribute.
  
CRIME \cite{redmond2002data}: includes demographic, economic and crime data about US communities with 1994 samples. We predict the number of violent crimes per $10^5$ population. As \cite{calders2013controlling}, we constructed a sensitive attribute based on Black population percentages.
  
LAW: corresponds to the Law School Admissions Councils National Longitudinal Bar Passage Study. The regression task consists in predicting students' GPA (normalized to the range $[0, 1]$) using race as the protected attribute.

\begin{figure}[htbp!]
    \centering
    \includegraphics[width=0.99\linewidth]{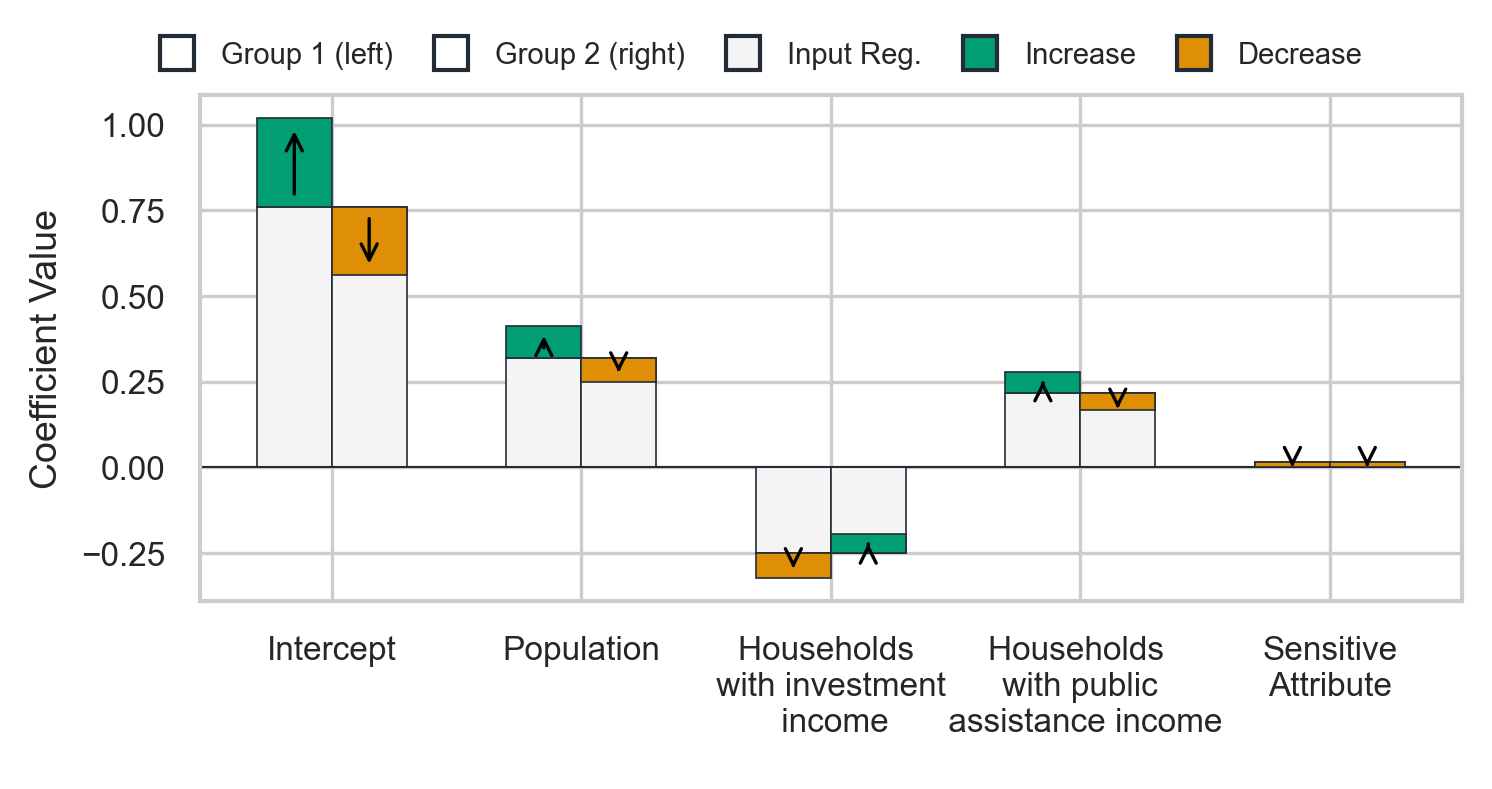}
    \caption{Analysis of coefficient shifts from the linear model to our fair model on the CRIME dataset.}
    \label{fig: Coefficients CRIME}
\end{figure}

\paragraph{Comparison w.r.t state-of-the-art.}

Experimental results (Table~\ref{tab:model_comparison}) demonstrate that our model effectively reduces unfairness while maintaining acceptable predictive performance. The Linear Model Unaware confirms that omitting the sensitive attribute fails to eliminate discriminatory patterns inherent in the data and does not prevent discriminatory outcomes. On the LAW dataset, unfairness is notably reduced between the Linear Model and the Linear Model Unaware, which explains the competitive performance of CS22 in this case, as this method is effective at mitigating direct bias; nevertheless, our model achieves even lower unfairness. Compared to the best-performing baselines, our approach achieves substantial unfairness reductions of 43\% on CRIME, 13\% on LAW, and 74\% on GOSSIS datasets. This improvement comes with a performance trade-off, showing approximately 15-20\% reduction in $GWR^2$ on CRIME while preserving competitive accuracy elsewhere.

\paragraph{Feature-level interpretation on CRIME Dataset}

While the direct bias is nullified (Fig.~\ref{fig: Coefficients CRIME}), the model mitigates indirect biases through group-specific coefficient adjustments. 

\section*{Conclusion}

We propose a closed-form solution for fair linear regression that enables exact control over the risk-fairness trade-off via the optimal predictor $f^*_\varepsilon$.  Building upon this Gaussian framework, we introduce a novel decomposition of unfairness into direct and indirect components, highlighting four distinct sources, including the previously overlooked \textbf{Indirect Structural Bias} arising from disparities in predictive variance.

Our results demonstrate that mean-based fairness alone is insufficient. By explicitly accounting for structural disparities, our method ensures fairness in both average predictions and predictive certainty across groups. The decomposition, along with the Group-Weighted $R^2$, provides actionable tools for diagnosing unfairness in linear models. While grounded in Gaussian assumptions, our approach shows strong empirical robustness on real-world data. Future work may extend these insights to non-linear models and broader fairness notions.

\bibliography{biblio}

\clearpage
\newpage

\appendix

\section{Notations and Background} \label{sub:Notations}
This section introduces the mathematical notations and fundamental concepts that serve as a foundation for the proofs and derivations that follow.
\subsection{Notation} Let $f$ be a function and $(\boldsymbol{X}, S) \in \mathcal{X} \times \mathcal{S} \subset \mathbb{R}^d \times \mathbb{N}$ a random pair, where $d$ is a positive integer and $S$ denotes a discrete sensitive attribute. Let $\mathcal{V}$ be the space of probability measures on a target space $\mathcal{Y} \subset \mathbb{R}$. We denote by $\nu_f \in \mathcal{V}$ the distribution of $f(\boldsymbol{X}, S)$, and by $\nu_{f \mid s} \in \mathcal{V}$ its conditional distribution given $S = s$. Let $F_{f \mid s}(u) := \mathbb{P}(f(\boldsymbol{X}, S) \leq u \mid S = s)$ be the cumulative distribution function (CDF) of $\nu_{f \mid s}$, and define the corresponding quantile function as $Q_{f \mid s}(v) := \inf \left\{ u \in \mathbb{R} : F_{f \mid s}(u) \geq v \right\}$.

\subsection{Background on Wasserstein distance and barycenter} \label{sub: Background Wasserstein}
This section reminds the concepts of Wasserstein distance and barycenter from one-dimensional optimal transport theory  \cite{santambrogio2015optimal}.
\begin{definition}[Wasserstein~\(\mathcal{W}_2\) Distance, see \cite{santambrogio2015optimal}]
For probability measures $\nu_0,\nu_1$ on $\mathbb{R}$, the squared 2–Wasserstein distance is
\[
\mathcal{W}_2(\nu_0,\nu_1)^2
\;=\;
\int_{0}^{1}
\bigl|Q_0(u)-Q_1(u)\bigr|^2
\,\mathrm{d}u.
\]
\end{definition}

Let $\{\nu_s : s\in\mathcal{S}\}$ be a finite family of measures on $\mathbb{R}$ with associated weights $\{p_s: s\in\mathcal{S}\}$.

\begin{definition}[Wasserstein Barycenter]
The Wasserstein barycenter $\nu^*$ is the unique minimizer
\[
\nu^*
\;=\;
\arg\min_{\nu\in\mathcal{P}_2(\mathbb{R})}
\sum_{s\in\mathcal{S}}p_s\,\mathcal{W}_2(\nu_s,\nu)^2,
\]
where $\mathcal{P}_2(\mathbb{R})$ denotes the set of probability measures on $\mathbb{R}$ with finite second moment.
\end{definition}



\section{Proof of Proposition~\ref{prop:optimal_fair_predictor}}
\label{appendix:proof_lemma_main}

This section provides a detailed proof for the closed-form expression of the optimal $\varepsilon$-DP predictor, $f^*_{\varepsilon}$, as stated in Proposition \ref{prop:optimal_fair_predictor}. Our derivation relies on the following foundational result concerning the characterization of the optimal exactly fair predictor, denoted $f^*_{DP}$.

\begin{theorem}[\cite{Chzhen_Denis_Hebiri_Oneto_Pontil20Wasser,gouic2020projection}]
    \label{thm:projection}
    Let $f^*$ be a predictor whose conditional distributions, $\nu_{f^*|s} := \mathrm{law}(f^*|S=s)$, have densities for all $s \in [M]$. The optimal predictor $f^*_{DP}$ that minimizes the risk $\mathcal{R}(f)$ while satisfying the Strong DP constraint is given by the composition of the average quantile function and the conditional CDF of $f^*$:
    \begin{equation*}
        f^*_{DP}(\boldsymbol{x},s) = \left(\sum_{s' \in [M]} p_{s'}F^{-1}_{f^*|s'} \right) \circ F_{f^*|s}(f^*(\boldsymbol{x},s)),
    \end{equation*}
    where $F_{f^*|s}$ is the CDF of $f^*$ conditional on $S=s$, and $F^{-1}_{f^*|s'}$ is its corresponding quantile function (inverse CDF).
\end{theorem}

\subsection{Proof of the optimality of $f^*_{DP}$ in our setting} 
\label{appx:sub_exact_fairness}
To apply the aforementioned theorem, we proceed in three steps: we first derive the conditional CDF of our Bayes predictor $f^*$, then its inverse, and finally, we assemble the expression for $f^*_{DP}$.

\paragraph{Deriving the Conditional CDF.}
Given our model, the Bayes predictor $f^*(\boldsymbol{X},S)$ conditional on $S=s$ is a linear transformation of a Gaussian random variable, and is therefore itself Gaussian. Its conditional mean and variance are:
\begin{align*}
    \mu_{f^*}^{(s)} &= \mathbb{E}[f^*(\boldsymbol{X},S)\mid S=s] = \langle \boldsymbol{\mu}^{(s)}, \boldsymbol{\beta^*} \rangle + \gamma^*s + \beta_0^*, \\
    (\sigma_{f^*}^{(s)})^2 &= \mathrm{Var}(f^*(\boldsymbol{X},S)\mid S=s) = (\boldsymbol{\beta^*})^\top\boldsymbol{\Sigma^{(s)}}\boldsymbol{\beta^*}.
\end{align*}
The conditional CDF, $F_{f^*|s}(t) = \mathbb{P}(f^*(\boldsymbol{X},S) \le t | S=s)$, is found by standardizing the variable:
\begin{align*}
    F_{f^*|s}(t) &= \mathbb{P}\left(\frac{f^*(\boldsymbol{X},S) - \mu_{f^*}^{(s)}}{\sigma_{f^*}^{(s)}} \le \frac{t - \mu_{f^*}^{(s)}}{\sigma_{f^*}^{(s)}} \bigg| S=s\right) \\
    &= \Phi\left(\frac{t - \mu_{f^*}^{(s)}}{\sigma_{f^*}^{(s)}}\right),
\end{align*}
where $\Phi(\cdot)$ is the CDF of the standard normal distribution.

\paragraph{Deriving the Conditional Quantile Function.}
The quantile function, $F^{-1}_{f^*\mid s}(p)$, is obtained by inverting the CDF expression. For a probability $p \in (0,1)$:
\begin{align*}
    p &= \Phi\left(\frac{F^{-1}_{f^*\mid s}(p) - \mu_{f^*}^{(s)}}{\sigma_{f^*}^{(s)}}\right) \\
    \implies \Phi^{-1}(p) &= \frac{F^{-1}_{f^*\mid s}(p) - \mu_{f^*}^{(s)}}{\sigma_{f^*}^{(s)}} \\
    \implies F^{-1}_{f^*\mid s}(p) &= \sigma_{f^*}^{(s)} \Phi^{-1}(p) + \mu_{f^*}^{(s)}.
\end{align*}

\paragraph{Assembling the Optimal DP Predictor.}
We now substitute these forms into the expression from Theorem \ref{thm:projection}. First, let's evaluate the inner part of the composition for a given input $(\boldsymbol{x},s)$:
\[
    F_{f^*\mid s}(f^*(\boldsymbol{x},s)) = \Phi\left(\frac{f^*(\boldsymbol{x},s) - \mu_{f^*}^{(s)}}{\sigma_{f^*}^{(s)}}\right).
\]
Now, we apply the averaged quantile function to this result. Let $p = F_{f^*\mid s}(f^*(\boldsymbol{x},s))$. The composition is:
\begin{align*}
    f^*_{DP}(\boldsymbol{x},s) &= \sum_{s' \in [M]} p_{s'} F^{-1}_{f^*\mid s'}(p) \\
    &= \sum_{s' \in [M]} p_{s'} \left( \sigma_{f^*}^{(s')} \Phi^{-1}(p) + \mu_{f^*}^{(s')} \right) \\
    &= \left(\sum_{s' \in [M]} p_{s'} \sigma_{f^*}^{(s')}\right) \Phi^{-1}(p) + \left(\sum_{s' \in [M]} p_{s'} \mu_{f^*}^{(s')}\right).
\end{align*}
Recognizing the definitions of the population-averaged moments, $\bar{\sigma}_{f^*} = \sum p_{s'}\sigma_{f^*}^{(s')}$ and $\bar{\mu}_{f^*} = \sum p_{s'}\mu_{f^*}^{(s')}$, we have:
$$
    f^*_{DP}(\boldsymbol{x},s) = \bar{\sigma}_{f^*} \Phi^{-1}(p) + \bar{\mu}_{f^*}.
$$
Finally, we substitute back the expression for $p$, where the $\Phi^{-1}$ and $\Phi$ functions cancel:
\begin{align*}
    f^*_{DP}(\boldsymbol{x},s) &= \bar{\sigma}_{f^*} \Phi^{-1}\left(\Phi\left(\frac{f^*(\boldsymbol{x},s) - \mu_{f^*}^{(s)}}{\sigma_{f^*}^{(s)}}\right)\right) + \bar{\mu}_{f^*} \\
    &= \bar{\sigma}_{f^*} \left(\frac{f^*(\boldsymbol{x},s) - \mu_{f^*}^{(s)}}{\sigma_{f^*}^{(s)}}\right) + \bar{\mu}_{f^*}.
\end{align*}
Using the definitions $f^*(\boldsymbol{x},s) = \langle \boldsymbol{x}, \boldsymbol{\beta^*} \rangle + \gamma^*s + \beta_0^*$ and $\mu_{f^*}^{(s)} = \langle \boldsymbol{\mu}^{(s)}, \boldsymbol{\beta^*} \rangle + \gamma^*s + \beta_0^*$, their difference is simply $f^*(\boldsymbol{x},s) - \mu_{f^*}^{(s)} = \langle \boldsymbol{x} - \boldsymbol{\mu}^{(s)}, \boldsymbol{\beta^*} \rangle$. This yields the final expression:
$$
    f^*_{DP}(\boldsymbol{x},s) = \bar{\sigma}_{f^*} \left( \frac{\langle \boldsymbol{x} - \boldsymbol{\mu}^{(s)}, \boldsymbol{\beta^*} \rangle}{\sigma_{f^*}^{(s)}} \right) + \bar{\mu}_{f^*},
$$

which proves the result stated in Proposition~\ref{prop:optimal_fair_predictor} for exact fairness. In the following, we prove the optimality of our expression in Approximate fairness.

\subsection{Proof of the optimality of $f^*_\varepsilon$ in our setting} \label{proof: optimality of the epsilon-RI predictor}

\paragraph{Proof Sketch.}
Our proof relies on the known general solution for the optimal $\varepsilon$-RI predictor from \cite{chzhen2022minimax}, which is the linear interpolation described in the following proposition.
\begin{proposition}[adapted from \cite{chzhen2022minimax}]
Assuming that $\{\nu^*_s\}_{s\in [K]}$ are non-atomic with finite second moments, then for all $\alpha \in [0,1]$ and all $(\boldsymbol{x},s) \in \mathbb{R}^p \times [K]$, the closed form solution of the minimization problem 
\begin{equation}\label{eq: minimization pb}
    \arg \min \left\{{\mathcal{R}}(f):{\mathcal{U}}(f)\leq \varepsilon \times {\mathcal{U}}(f^*)\right\}
\end{equation}
is given by:
\begin{align*}
    f^*_\varepsilon (\boldsymbol{x},s) &= \sqrt{\varepsilon}f^*(\boldsymbol{x},s)\\&+(1-\sqrt{\varepsilon})\sum_{s'=1}^K p_{s'}F^{-1}_{\nu^*_{s'}}\circ F_{\nu^*_s}\circ f^*(\boldsymbol{x},s) \nonumber\\
   &= \sqrt{\varepsilon}f^*(\boldsymbol{x},s)+(1-\sqrt{\varepsilon})f^*_{DP}(\boldsymbol{x},s)
\end{align*}
where $f^*_{DP}(\boldsymbol{x},s)$ represents the demographic parity optimal predictor.
\end{proposition}
We will then substitute the explicit expressions for our Bayes predictor $f^*$ and optimal DP predictor $f^*_{DP}$ and show that the resulting expression simplifies to the form stated in the proposition~\ref{prop:optimal_fair_predictor}.

\paragraph{Expressing Predictors in a Common Form.}
To simplify the algebra, we first express both $f^*$ and $f^*_{DP}$ in terms of a common standardized score. Let the group-standardized score for an input $\boldsymbol{x}$ be:
\[
    z^{(s)}(\boldsymbol{x}) = \frac{\langle \boldsymbol{x} - \boldsymbol{\mu}^{(s)}, \boldsymbol{\beta^*} \rangle}{\sigma_{f^*}^{(s)}}.
\]
By construction, the random variable $z^{(s)}(\boldsymbol{X})$ has a conditional mean of 0 and a conditional standard deviation of 1 for group $s$.

Using the results from the proof in \ref{appx:sub_exact_fairness}, we can write both predictors as linear transformations of this standardized score:
\begin{itemize}
    \item The Bayes predictor $f^*$ can be rewritten as:
    \begin{align*}
        f^*(\boldsymbol{x},s) &= \langle \boldsymbol{x}, \boldsymbol{\beta^*} \rangle + \gamma^*s + \beta_0^* \\ 
        & = \langle \boldsymbol{x} - \boldsymbol{\mu}^{(s)}, \boldsymbol{\beta^*} \rangle + \langle \boldsymbol{\mu}^{(s)}, \boldsymbol{\beta^*} \rangle + \gamma^*s + \beta_0^* \\
        &= \sigma_{f^*}^{(s)} \cdot z^{(s)}(\boldsymbol{x}) + \mu_{f^*}^{(s)}.
    \end{align*}
    \item The optimal DP predictor $f^*_{DP}$ from proof in \ref{appx:sub_exact_fairness} is:
    $$
    f^*_{DP}(\boldsymbol{x},s) = \bar{\sigma}_{f^*} \cdot z^{(s)}(\boldsymbol{x}) + \bar{\mu}_{f^*}.
    $$
\end{itemize}

\paragraph{Substituting and Simplifying the Interpolation.}
Now we substitute these two expressions back into the interpolation formula:
\begin{align*}
    f^*_\varepsilon(\boldsymbol{x},s) &= (1-\sqrt{\varepsilon}) \left( \bar{\sigma}_{f^*} \cdot z^{(s)}(\boldsymbol{x}) + \bar{\mu}_{f^*} \right) \\ &+ \sqrt{\varepsilon} \left( \sigma_{f^*}^{(s)} \cdot z^{(s)}(\boldsymbol{x}) + \mu_{f^*}^{(s)} \right).
\end{align*}
We can now collect the terms multiplying the standardized score $z^{(s)}(\boldsymbol{x})$ and the constant terms separately:
\begin{align*}
    f^*_\varepsilon(\boldsymbol{x},s) &= \left[ (1-\sqrt{\varepsilon})\bar{\sigma}_{f^*} + \sqrt{\varepsilon}\sigma_{f^*}^{(s)} \right] \cdot z^{(s)}(\boldsymbol{x}) \\ &+ \left[ (1-\sqrt{\varepsilon})\bar{\mu}_{f^*} + \sqrt{\varepsilon}\mu_{f^*}^{(s)} \right].
\end{align*}

\paragraph{Identifying the Final Form.}
We recognize the expressions in the brackets as the definitions of the effective standard deviation $\sigma_{\varepsilon}^{(s)}$ and mean $\mu_{\varepsilon}^{(s)}$ from the proposition statement:
\begin{align*}
    \sigma_{\varepsilon}^{(s)} &= (1-\sqrt{\varepsilon}) \bar{\sigma}_{f^*} + \sqrt{\varepsilon}\sigma_{f^*}^{(s)} \\
    \mu_{\varepsilon}^{(s)} &= (1-\sqrt{\varepsilon}) \bar{\mu}_{f^*} + \sqrt{\varepsilon}\mu_{f^*}^{(s)}.
\end{align*}
Substituting these definitions back, we obtain:
$$
f^*_\varepsilon(\boldsymbol{x},s) = \sigma_{\varepsilon}^{(s)} \cdot z^{(s)}(\boldsymbol{x}) + \mu_{\varepsilon}^{(s)}.
$$
Finally, substituting the definition of $z^{(s)}(\boldsymbol{x})$ gives the exact expression from the proposition:
\[
    f^*_\varepsilon(\boldsymbol{x},s) = \sigma_{\varepsilon}^{(s)} \left( \frac{\langle \boldsymbol{x} - \boldsymbol{\mu}^{(s)}, \boldsymbol{\beta^*} \rangle}{\sigma_{f^*}^{(s)}} \right) + \mu_{\varepsilon}^{(s)}.
\]
This completes the proof.




\section{Proof of Proposition \ref{prop:bias_decomp} and Corollary~\ref{cor:unfairness_epsilon}}
\label{sub:proof:bias_decomp}

This section provides a detailed proof for the characterization of the Bias decomposition. We begin by stating the foundational results from optimal transport theory that underpin our analysis.

\subsection{Preliminaries for the computation of the unfairness in our framework}
\label{appendix:preliminaries}

Our unfairness measure is based on the Wasserstein-2 distance. Its computation for Gaussian distributions relies on two key results.

First, the squared Wasserstein-2 distance between two one-dimensional Gaussian distributions has a simple closed form.
\begin{lemma}[From \cite{frechet1957distance}]
\label{lemma: frechet}
For any means $\mu_1, \mu_2 \in \mathbb{R}$ and standard deviations $\sigma_1, \sigma_2 \in \mathbb{R}^+$, the squared $\mathcal{W}_2$ distance between the corresponding Gaussian distributions is:
\[
\mathcal{W}_2^2(\mathcal{N}(\mu_1,\sigma_1^2), \mathcal{N}(\mu_2,\sigma_2^2)) = (\mu_1-\mu_2)^2 + (\sigma_1-\sigma_2)^2.
\]
\end{lemma}

Second, the Wasserstein barycenter of a set of Gaussian distributions is also a Gaussian, whose moments are the weighted averages of the input moments.
\begin{lemma}[From \cite{agueh2011barycenters}]
\label{lemma: barycenter for gaussian}
Let $(\mathcal{N}(\mu_s, \sigma_s^2))_{s \in [M]}$ be a set of Gaussian distributions and let $(p_s)_{s \in [M]}$ be a probability vector. The unique Wasserstein barycenter that solves the minimization problem in Eq.~\eqref{eq:unfairness} is the Gaussian distribution $\mathcal{N}(\bar{\mu}, \bar{\sigma}^2)$, where:
\[
\bar{\mu} = \sum_{s=1}^{M} p_{s} \mu_s \quad \text{and} \quad \bar{\sigma} = \sum_{s=1}^{M} p_{s} \sigma_s.
\]
\end{lemma}

\subsection{Proof of Proposition \ref{prop:bias_decomp}}

\begin{proof}
The proof proceeds in three steps. First, we simplify the definition of $\mathcal{U}(f)$ under our Gaussian assumption. Second, we decompose the first-moment term. Third, we decompose the second-moment term.

\paragraph{Step 1: From Wasserstein Distance to a Variance Decomposition.}
The unfairness measure is defined as:
\[
    \mathcal{U}(f) = \min_{\nu \in \mathcal{P}_2(\mathbb{R})}\sum_{s=1}^M p_s W_2^2(\nu_{f\mid s},\nu).
\]
Under our modeling assumption, the conditional distribution $\nu_{f\mid s}$ is Gaussian with mean $\mu_f^{(s)} = \mathbb{E}[f|S=s]$ and standard deviation $\sigma_f^{(s)} = \sqrt{\mathrm{Var}(f\mid S=s)}$. Using Lemma~\ref{lemma: barycenter for gaussian}, the Wasserstein barycenter $\nu^*$ is also a Gaussian, say $\mathcal{N}(\mu_\nu, \sigma_\nu^2)$. Substituting this into the definition of $\mathcal{U}(f)$ gives:
\[
    \mathcal{U}(f) = \min_{\mu_\nu, \sigma_\nu} \sum_{s=1}^M p_s \left[ (\mu_f^{(s)} - \mu_\nu)^2 + (\sigma_f^{(s)} - \sigma_\nu)^2 \right].
\]
This optimization problem decouples. The optimal $\mu_\nu^*$ is the population average mean, $\mathbb{E}_S[\mu_f^{(S)}]$, and the optimal $\sigma_\nu^*$ is the population average standard deviation, $\mathbb{E}_S[\sigma_f^{(S)}]$. Plugging these back and using Lemma~\ref{lemma: frechet} gives the exact decomposition:
\begin{align*}
    \mathcal{U}(f) &= \sum_{s=1}^M p_s (\mu_f^{(s)} - \mathbb{E}[\mu_f^{(S)}])^2 + \sum_{s=1}^M p_s (\sigma_f^{(s)} - \mathbb{E}[\sigma_f^{(S)}])^2 \\
    &= \mathrm{Var}(\mathbb{E}[f\mid S]) + \mathrm{Var}(\sqrt{\mathrm{Var}(f\mid S)}).
\end{align*}

\paragraph{Step 2: Decomposing the First-Moment Disparity.}
The FMD, $\mathrm{Var}(\mathbb{E}[f\mid S])$, for $f(\boldsymbol{X}, S) = \langle \boldsymbol{X}, \boldsymbol{\beta} \rangle + \gamma S + \beta_0$ is derived as:
\begin{multline*}
    \mathrm{Var}(\mathbb{E}[f\mid S]) \\ = \underbrace{\gamma^2 \mathrm{Var}(S)}_{\text{Direct Mean Bias}} + \underbrace{\mathrm{Var}(\langle \boldsymbol{\mu}^{(S)}, \boldsymbol{\beta} \rangle)}_{\text{Indirect Mean Bias}} + \underbrace{2\gamma \mathrm{Cov}(S, \langle \boldsymbol{\mu}^{(S)}, \boldsymbol{\beta} \rangle)}_{\text{Interaction}}.
\end{multline*}

\paragraph{Step 3: Decomposing the Second-Moment Disparity.}
The SMD is $\mathrm{Var}(\sqrt{\mathrm{Var}(f\mid S)})$. The conditional variance of $f$ is $\mathrm{Var}(f\mid S=s) = \boldsymbol{\beta}^\top \boldsymbol{\Sigma}^{(s)} \boldsymbol{\beta}$. Therefore, the conditional standard deviation is $\sigma_f^{(s)} = \sqrt{\boldsymbol{\beta}^\top \boldsymbol{\Sigma}^{(s)} \boldsymbol{\beta}}$. The SMD is the variance of this quantity:
\[
    \mathrm{Var}(\sqrt{\mathrm{Var}(f\mid S)}) = \mathrm{Var}\left(\sqrt{\boldsymbol{\beta}^\top \boldsymbol{\Sigma}^{(S)} \boldsymbol{\beta}}\right).
\]
This term corresponds to the \textbf{Indirect Structural Bias}.

Combining the results from Step 2 and Step 3 gives the full four-term decomposition stated in the proposition.
\end{proof}

In the following, we provide the detailed derivations for the unfairness of the Bayes predictor ($f^*$) and our optimal DP predictor ($f^*_{DP}$), applying the general result from Appendix~\ref{appendix:preliminaries} that $\mathcal{U}(f) = \mathrm{Var}(\mathbb{E}[f|S]) + \mathrm{Var}(\sqrt{\mathrm{Var}(f|S)})$.

\paragraph{Unfairness of the Bayes Predictor, $\mathcal{U}(f^*)$.} We apply the general unfairness formula to the Bayes predictor $f^* \in \mathcal{F}_{\text{linear}}$, which has coefficients $(\boldsymbol{\beta}^*, \gamma^*, \beta_0^*)$. The total unfairness naturally gives the FMD and SMD decomposition with:
\begin{enumerate}
    \item \textbf{First-Moment Disparity:} This is the variance of the conditional mean, $\mathrm{Var}(\mu_{f^*}^{(S)})$:
    \begin{multline*}
        \mathrm{Var}(\mu_{f^*}^{(S)}) = \\ \mathrm{Var}(\langle \boldsymbol{\mu}^{(S)}, \boldsymbol{\beta^*} \rangle) + \mathrm{Var}(\gamma^*S) + 2\mathrm{Cov}(\langle \boldsymbol{\mu}^{(S)}, \boldsymbol{\beta^*} \rangle, \gamma^*S).
    \end{multline*}
    This corresponds exactly to the sum of the Indirect Mean Bias, Direct Mean Bias, and Interaction terms for the Bayes predictor.

    \item \textbf{Second-Moment Disparity:} This is the variance of the conditional standard deviation, $\mathrm{Var}(\sigma_{f^*}^{(S)})$:
    $$
        \mathrm{Var}(\sigma_{f^*}^{(S)}) = \mathrm{Var}\left(\sqrt{(\boldsymbol{\beta^*})^\top\boldsymbol{\Sigma}^{(S)}\boldsymbol{\beta^*}}\right).
    $$
    This is the Indirect Structural Bias for the Bayes predictor.
\end{enumerate}

\paragraph{Unfairness of the Optimal DP Predictor, $\mathcal{U}(f^*_{DP})$.}
We now prove that our optimal fair predictor, $f^*_{DP}$, has zero unfairness. We start with the definition of $f^*_{DP}$ from Proposition~\ref{prop:optimal_fair_predictor} with $\varepsilon=0$:
$$
    f^*_{DP}(\boldsymbol{x},s) := \bar{\sigma}_{f^*} \left( \frac{\langle \boldsymbol{x} - \boldsymbol{\mu}^{(s)}, \boldsymbol{\beta^*} \rangle}{\sigma_{f^*}^{(s)}} \right) + \bar{\mu}_{f^*}.
$$
To compute its unfairness, we must find its conditional moments given $S=s$.

\begin{enumerate}
    \item \textbf{Conditional Mean of $f^*_{DP}$:}
    \begin{align*}
        \mathbb{E}[f^*_{DP}(\boldsymbol{X}&,S)|S=s]  \\ &= \mathbb{E}\left[ \bar{\sigma}_{f^*} \left( \frac{\langle \boldsymbol{X} - \boldsymbol{\mu}^{(s)}, \boldsymbol{\beta^*} \rangle}{\sigma_{f^*}^{(s)}} \right) + \bar{\mu}_{f^*} \bigg| S=s \right] \\
        &= \frac{\bar{\sigma}_{f^*}}{\sigma_{f^*}^{(s)}} \mathbb{E}[\langle \boldsymbol{X} - \boldsymbol{\mu}^{(s)}, \boldsymbol{\beta^*} \rangle | S=s] + \bar{\mu}_{f^*} \\
        &= \frac{\bar{\sigma}_{f^*}}{\sigma_{f^*}^{(s)}} \langle \mathbb{E}[\boldsymbol{X}|S=s] - \boldsymbol{\mu}^{(s)}, \boldsymbol{\beta^*} \rangle + \bar{\mu}_{f^*} \\
        &= \frac{\bar{\sigma}_{f^*}}{\sigma_{f^*}^{(s)}} \langle \boldsymbol{\mu}^{(s)} - \boldsymbol{\mu}^{(s)}, \boldsymbol{\beta^*} \rangle + \bar{\mu}_{f^*} \\
        &= 0 + \bar{\mu}_{f^*}.
    \end{align*}
    The conditional mean, $\mathbb{E}[f^*_{DP}|S=s]$, is equal to the constant $\bar{\mu}_{f^*}$ for all groups $s$.

    \item \textbf{Conditional Variance of $f^*_{DP}$:}
    \begin{align*}
        \mathrm{Var}(f^*_{DP}(\boldsymbol{X}&,S)|S=s) \\
        &= \mathrm{Var}\left( \frac{\bar{\sigma}_{f^*}}{\sigma_{f^*}^{(s)}} \langle \boldsymbol{X} - \boldsymbol{\mu}^{(s)}, \boldsymbol{\beta^*} \rangle + \bar{\mu}_{f^*} \bigg| S=s \right) \\
        &= \left(\frac{\bar{\sigma}_{f^*}}{\sigma_{f^*}^{(s)}}\right)^2 \mathrm{Var}(\langle \boldsymbol{X}, \boldsymbol{\beta^*} \rangle | S=s) \\
        &= \left(\frac{\bar{\sigma}_{f^*}}{\sigma_{f^*}^{(s)}}\right)^2 ((\sigma_{f^*}^{(s)})^2) \\
        &= (\bar{\sigma}_{f^*})^2.
    \end{align*}
    The conditional variance, $\mathrm{Var}(f^*_{DP}|S=s)$, is equal to the constant $(\bar{\sigma}_{f^*})^2$ for all groups $s$. The conditional standard deviation is therefore also constant: $\sqrt{\mathrm{Var}(f^*_{DP}|S=s)} = \bar{\sigma}_{f^*}$.
\end{enumerate}
Since both the conditional mean and the conditional standard deviation of $f^*_{DP}$ are constant across all groups $s$, their variance with respect to $S$ is zero.
Therefore, the total unfairness is $\mathcal{U}(f^*_{DP}) = 0 + 0 = 0$.

\subsection{Proof of the Corollary~\ref{cor:unfairness_epsilon}: Unfairness of $\varepsilon$-RI predictor}

\begin{proof}
The corollary states that the total unfairness of the optimal $\varepsilon$-fair predictor, $\mathcal{U}(f^*_\varepsilon)$, is exactly $\varepsilon$ times the unfairness of the original Bayes predictor, $\mathcal{U}(f^*)$.

We recall the definition of the optimal $\varepsilon$-RI predictor as the geodesic interpolation between the optimal fair predictor ($f^*_{DP}$) and the Bayes predictor ($f^*$):
$$
f^*_\varepsilon = (1-\sqrt{\varepsilon})f^*_{DP} + \sqrt{\varepsilon}f^*.
$$
We will compute the total unfairness of $f^*_\varepsilon$ using the exact Wasserstein decomposition derived in the proof of Proposition \ref{prop:bias_decomp}:
$$
\mathcal{U}(f^*_\varepsilon) = \mathrm{Var}(\mathbb{E}[f^*_\varepsilon|S]) + \mathrm{Var}(\sqrt{\mathrm{Var}(f^*_\varepsilon|S)}).
$$
Let us compute each of the two variance terms separately.

We first compute the conditional expectation of $f^*_\varepsilon$ given $S=s$. By linearity of expectation:
\begin{align*}
    \mathbb{E}[f^*_\varepsilon|S=s] &= \mathbb{E}[(1-\sqrt{\varepsilon})f^*_{DP} + \sqrt{\varepsilon}f^*|S=s] \\
    &= (1-\sqrt{\varepsilon})\mathbb{E}[f^*_{DP}|S=s] + \sqrt{\varepsilon}\mathbb{E}[f^*|S=s].
\end{align*}
From the analysis of $f^*_{DP}$ in the proof of our main Proposition~\ref{prop:optimal_fair_predictor}, we know that its conditional mean is the constant population average, $\mathbb{E}[f^*_{DP}|S=s] = \bar{\mu}_{f^*}$. The conditional mean of the Bayes predictor is simply $\mu_{f^*}^{(s)}$. Substituting these in gives:
$$
    \mathbb{E}[f^*_\varepsilon|S=s] = (1-\sqrt{\varepsilon})\bar{\mu}_{f^*} + \sqrt{\varepsilon}\mu_{f^*}^{(s)}.
$$
Now, we compute the variance of this expression with respect to the random variable $S$. Since $\bar{\mu}_{f^*}$ is a constant, it does not contribute to the variance:
\begin{align*}
    \mathrm{Var}(\mathbb{E}[f^*_\varepsilon|S]) &= \mathrm{Var}((1-\sqrt{\varepsilon})\bar{\mu}_{f^*} + \sqrt{\varepsilon}\mu_{f^*}^{(S)}) \\
    &= \mathrm{Var}(\sqrt{\varepsilon}\mu_{f^*}^{(S)}) \\
    & = \varepsilon \cdot \mathrm{Var}(\mathbb{E}[f^*|S]).
\end{align*}

\paragraph{Decomposing the Second-Moment Disparity of $f^*_\varepsilon$.}

Next, we compute the conditional variance, $\mathrm{Var}(f^*_\varepsilon|S=s)$. As shown in the proof of Proposition \ref{prop:optimal_fair_predictor}, the predictor $f^*_\varepsilon$ can be expressed in terms of a standardized score, $z^{(s)}(\boldsymbol{x}) = \langle \boldsymbol{x} - \boldsymbol{\mu}^{(s)}, \boldsymbol{\beta^*} \rangle / \sigma_{f^*}^{(s)}$, which has a conditional variance of 1. The predictor is:
$$
f^*_\varepsilon(\boldsymbol{x},s) = \sigma_{\varepsilon}^{(s)} \cdot z^{(s)}(\boldsymbol{x}) + \mu_{\varepsilon}^{(s)}.
$$
Since $\sigma_{\varepsilon}^{(s)}$ and $\mu_{\varepsilon}^{(s)}$ are constant given $S=s$, the conditional variance is:
\begin{align*}
    \mathrm{Var}(f^*_\varepsilon|S=s) &= \mathrm{Var}(\sigma_{\varepsilon}^{(s)} \cdot z^{(s)}(\boldsymbol{X})|S=s) \\
    & = (\sigma_{\varepsilon}^{(s)})^2 \mathrm{Var}(z^{(s)}(\boldsymbol{X})|S=s) \\
    &= (\sigma_{\varepsilon}^{(s)})^2.
\end{align*}
The conditional standard deviation is therefore simply $\sigma_{\varepsilon}^{(s)}$:
$$
\sqrt{\mathrm{Var}(f^*_\varepsilon|S=s)} = \sigma_{\varepsilon}^{(s)} = (1-\sqrt{\varepsilon})\bar{\sigma}_{f^*} + \sqrt{\varepsilon}\sigma_{f^*}^{(s)}.
$$
Since $\bar{\sigma}_{f^*}$ is a constant, it does not contribute to the variance:
\begin{align*}
    \mathrm{Var}(\sqrt{\mathrm{Var}(f^*_\varepsilon|S)}) &= \mathrm{Var}((1-\sqrt{\varepsilon})\bar{\sigma}_{f^*} + \sqrt{\varepsilon}\sigma_{f^*}^{(S)}) \\
    &= \mathrm{Var}(\sqrt{\varepsilon}\sigma_{f^*}^{(S)}) \\
    &= (\sqrt{\varepsilon})^2 \mathrm{Var}(\sigma_{f^*}^{(S)}) \\
    &= \varepsilon \cdot \mathrm{Var}(\sqrt{\mathrm{Var}(f^*|S)}).
\end{align*}
\paragraph{Assembling the Final Result.}
Finally, we sum the two decomposed disparity terms:
\begin{align*}
    \mathcal{U}(f^*_\varepsilon) &= \mathrm{Var}(\mathbb{E}[f^*_\varepsilon|S]) + \mathrm{Var}(\sqrt{\mathrm{Var}(f^*_\varepsilon|S)}) \\
    &= \varepsilon \cdot \mathrm{Var}(\mathbb{E}[f^*|S]) + \varepsilon \cdot \mathrm{Var}(\sqrt{\mathrm{Var}(f^*|S)}) \\
    &= \varepsilon \cdot \left( \mathrm{Var}(\mathbb{E}[f^*|S]) + \mathrm{Var}(\sqrt{\mathrm{Var}(f^*|S)}) \right) \\
    &= \varepsilon \cdot \mathcal{U}(f^*).
\end{align*}
This completes the proof.

\end{proof}

\section{Proof of Proposition~\ref{prop:feature_decomp_diag} : Additive Feature-Level Decomposition}
\label{appendix:feature_level_proof}
This appendix provides the full derivation for the additive approximation of the feature-level decomposition of indirect unfairness, as stated in Proposition~\ref{prop:feature_decomp_diag}.

\subsection{The General Decomposition with Interactional Terms}
Our goal is to decompose the indirect unfairness of a predictor $f \in \mathcal{F}_{\text{linear}}$ with coefficients $(\boldsymbol{\beta}, \gamma)$. We begin with the exact expression for indirect unfairness:
\begin{align*}
    \mathcal{U}_{\mathrm{indirect}}(f) &= \mathrm{Var}(\langle \boldsymbol{\mu}^{(S)}, \boldsymbol{\beta} \rangle)\\
    &+ 2\gamma \mathrm{Cov}(S, \langle \boldsymbol{\mu}^{(S)}, \boldsymbol{\beta} \rangle) \\
    &+ \mathrm{Var}\left(\sqrt{\boldsymbol{\beta}^\top\boldsymbol{\Sigma}^{(S)}\boldsymbol{\beta}}\right).
\end{align*}
Even under the simplifying assumption of uncorrelated features (diagonal $\Sigma^{(s)}$), a full decomposition reveals the presence of cross-terms that capture the statistical relationships between the group-level properties of different features.

\paragraph{Decomposition of Mean-Based Terms.}
The mean-based terms decompose as follows:
\begin{align*}
    \mathrm{Var}(\langle \boldsymbol{\mu}^{(S)}, \boldsymbol{\beta} \rangle) &= \sum_{j=1}^d (\beta_j)^2 \mathrm{Var}(\mu_j^{(S)}) \\
    &+ 2\sum_{j<k} \beta_j \beta_k \mathrm{Cov}(\mu_j^{(S)}, \mu_k^{(S)}), \\
    2\gamma \mathrm{Cov}(S, \langle \boldsymbol{\mu}^{(S)}, \boldsymbol{\beta} \rangle) &= \sum_{j=1}^d 2\gamma \beta_j \mathrm{Cov}(S, \mu_j^{(S)}).
\end{align*}
The term $\mathrm{Cov}(\mu_j^{(S)}, \mu_k^{(S)})$ represents the \textbf{compounding of mean biases}.

\paragraph{Decomposition of the Structural Term (via Linearization).}
The structural bias term, $\mathrm{Var}(\sqrt{\boldsymbol{\beta}^\top\boldsymbol{\Sigma}^{(S)}\boldsymbol{\beta}})$, is non-linear. To decompose it, we first apply a first-order Taylor expansion. Let $V(S) = \boldsymbol{\beta}^\top\boldsymbol{\Sigma}^{(S)}\boldsymbol{\beta}$ be the conditional score variance. We linearize the function $g(v)=\sqrt{v}$ around $\bar{V} = \mathbb{E}[V(S)]$:
$$
    \sqrt{V(S)} \approx \sqrt{\bar{V}} + \frac{1}{2\sqrt{\bar{V}}}(V(S) - \bar{V}).
$$
Taking the variance of this linear approximation yields:
$$
    \mathrm{Var}(\sqrt{V(S)}) \approx \frac{1}{4\bar{V}} \mathrm{Var}(V(S)).
$$
Under the diagonal $\Sigma^{(s)}$ assumption, 
$$
V(S) = \sum_{j=1}^d \beta_j^2 (\sigma_j^{(S)})^2\enspace.
$$
The variance of this sum is:
\begin{multline*}
    \mathrm{Var}(V(S)) \\= \sum_{j=1}^d \beta_j^4 \mathrm{Var}((\sigma_j^{(S)})^2) + 2 \sum_{j<k} \beta_j^2 \beta_k^2 \mathrm{Cov}((\sigma_j^{(S)})^2, (\sigma_k^{(S)})^2).
\end{multline*}
The term $\mathrm{Cov}((\sigma_j^{(S)})^2, (\sigma_k^{(S)})^2)$ represents the \textbf{compounding of structural biases}.

\subsection{Proof of the Additive Approximation (Proposition \ref{prop:feature_decomp_diag})}
\begin{proof}
By combining the full decompositions above, we can express the exact indirect unfairness as a sum of two components: the primary (approximated) contributions from each feature and the interactional unfairness from pairs of features.
\begin{align*}
    \mathcal{U}_{\mathrm{indirect}}(f) &\approx \sum_{j=1}^d \bigg( (\beta_j)^2 \mathrm{Var}(\mu_j^{(S)}) + \\ & \quad\quad \frac{1}{4\bar{V}} (\beta_j)^4 \mathrm{Var}((\sigma_j^{(S)})^2) + 2\gamma \beta_j \mathrm{Cov}(S, \mu_j^{(S)}) \bigg) \\
    &+ 2 \sum_{j<k} \bigg[ \beta_j \beta_k \mathrm{Cov}(\mu_j^{(S)}, \mu_k^{(S)}) + \\ & \quad\quad 
    \frac{1}{4\bar{V}} \beta_j^2 \beta_k^2 \mathrm{Cov}((\sigma_j^{(S)})^2, (\sigma_k^{(S)})^2) \bigg].
\end{align*}
We define the first sum as the sum of the `Approximated' primary feature contributions, $\sum_j \mathcal{U}^{\text{approx}}_j(f)$.

The proposition provides an additive approximation of the total indirect unfairness. This approximation is formally derived by neglecting the second term, the \textit{Interactional Unfairness}. This is an approximation that can be highly accurate under the common scenario where the primary contributions from individual features are significantly larger than the second-order interactional effects.
Therefore, we have:
\begin{align*}
    \mathcal{U}_{\mathrm{indirect}}(f) &\approx \sum_{j=1}^d \mathcal{U}^{\text{approx}}_j(f) \\
    &= \sum_{j=1}^d \bigg( (\beta_j)^2 \mathrm{Var}(\mu_j^{(S)}) + \frac{1}{4\bar{V}} (\beta_j)^4 \mathrm{Var}((\sigma_j^{(S)})^2)\\ & \quad\quad + 2\gamma \beta_j \mathrm{Cov}(S, \mu_j^{(S)}) \bigg).
\end{align*}
This completes the proof.
\end{proof}

\section{Auditing Model Adequacy with the $GWR^2$}
\label{app:gwr2_details}

We prove the precise statistical conditions under which the global $R^2$ and our proposed $GWR^2$ are equivalent, demonstrating that their divergence is a direct consequence of underlying group-level disparities in the data.
\begin{proposition}[Characterization of the $GWR^2$]
\label{prop: characterization of GWR2}
Let $f(\boldsymbol{X}, S)$ be a linear model of the form
$$
f(\boldsymbol{X}, S) = \langle \boldsymbol{X}, \boldsymbol{\beta} \rangle +\gamma S+\beta_0 \enspace.
$$
We also define the global coefficient of determination by
$$
R^2_{\mathrm{global}} := 1 - \frac{\mathrm{Var}(Y - f(\boldsymbol{X}, S))}{\mathrm{Var}(Y)}\enspace .
$$
Then, we have $R^2_{\mathrm{global}} = GWR^2$ iff all three are verified:
\begin{enumerate}
  \item[(i)] \textbf{(Unaware model)} The model does not include the sensitive attribute as a predictor. This means we set $\gamma = 0$.
  \item[(ii)] \textbf{(No residual association)} After accounting for $\boldsymbol{X}$, the attribute $S$ has no remaining linear association with the outcome $Y$. This means the true parameter $\gamma^* = 0$.
  \item[(iii)] \textbf{(Feature independence)} The features $\boldsymbol{X}$ are independent of the sensitive attribute $S$. 
\end{enumerate}
\end{proposition}
Proof can be found in Appendix~\ref{sub: proof of prop characterization of GWR2}.

\paragraph{Decomposition of the $R^2$ Discrepancy}

When the strict conditions for equality between the global and group-weighted $R^2$ metrics are not met, a discrepancy arises. The following proposition provides an exact analytical expression for this discrepancy under the simplifying assumption of homoscedasticity, revealing that the gap is driven by how a \textit{fair-unaware} model accounts for the between-group variance present in the data.

\begin{proposition}[Decomposition of the $R^2$ Gap] \label{prop: decomposition of the R2 Gap}
Let $f(\boldsymbol{X})$ be a fair-unaware predictor. For simplicity sake we assume (only in this proposition) that the conditional variances of both the outcome and the residuals are constant across all groups $s \in \mathcal{S}$ (homoscedasticity):
\begin{itemize}
    \item Within-group outcome variance: 
    $$
    \mathrm{Var}(Y \mid S=s) = W_Y, \quad \text{for all } s \in\mathcal{S}
    $$
    \item Within-group residual variance: 
    $$
    \mathrm{Var}(Y - f(\boldsymbol{X}) \mid S=s) = W_R, \quad \text{for all } s \in\mathcal{S}
    $$
\end{itemize}
Let $B_Y := \mathrm{Var}(\mathbb{E}[Y \mid S])$ denote the between-group variance of the outcome, and let $B_R := \mathrm{Var}(\mathbb{E}[Y - f(\boldsymbol{X}) \mid S])$ denote the between-group variance of the model's residuals.

Then the difference between the group-weighted and global R-squared metrics is given by:
\begin{equation} \label{eq:R2_gap}
    GWR^2 - R^2_{\mathrm{global}} = \frac{W_Y B_R - W_R B_Y}{W_Y (W_Y + B_Y)}
\end{equation}
\end{proposition}
Proof can be found in Appendix~\ref{sub: proof of decomposition of the R2 Gap}.

\paragraph{Implication}
The decomposition in \eqref{eq:R2_gap} provides a clear interpretation of the discrepancy. Specifically, we show that $GWR^2 > R^2_{\mathrm{global}}$ if and only if $W_Y B_R > W_R B_Y$, which can be rewritten as:
$$
\frac{B_R}{W_R} > \frac{B_Y}{W_Y}\enspace.
$$
This inequality compares the ratio of between-group to within-group variance for the residuals ($R$) and for the original outcome ($Y$). When a fair-unaware model $f$ fails to explain the between-group variance ($B_Y$) that exists in the outcome, this unexplained variance is transferred to the residuals, inflating $B_R$. This leads to a situation where the model appears to perform worse from a global perspective than it does on average within the groups. The gap between the two R-squared metrics is therefore a direct signal of how well the model accounts for the group-level structural differences present in the data.

\subsection{Proof of Proposition \ref{prop: characterization of GWR2}} \label{sub: proof of prop characterization of GWR2}

\begin{proof}
The equality $R^2_{\mathrm{global}} = GWR^2$ holds if and only if their fractional parts are equal:
\begin{equation}
    \frac{\mathrm{Var}(Y - f)}{\mathrm{Var}(Y)} = \sum_s p_{s} \frac{\mathrm{Var}(Y - f(\cdot, s) \mid S=s)}{\mathrm{Var}(Y \mid S=s)} \tag{*}
\end{equation}

\paragraph{Part 1: ($\Leftarrow$)}
Assume conditions (i), (ii), and (iii) are all true. Our goal is to show that equation (*) holds.

\medskip
\textit{Step A: Equality of denominators.} We first show that $\mathrm{Var}(Y) = \mathrm{Var}(Y \mid S=s)$ for all $s$.
By the Law of Total Variance, 
$$
\mathrm{Var}(Y) = \mathbb{E}[\mathrm{Var}(Y\mid S)] + \mathrm{Var}(\mathbb{E}[Y\mid S])\enspace .
$$
From Eq.~\eqref{eq:our_model} and using condition (ii) ($\gamma^*=0$), we have $Y = \langle \boldsymbol{X}, \boldsymbol{\beta^*} \rangle + \beta_0^* + \zeta$.
\begin{itemize}
    \item The conditional variance is 
    $$
    \mathrm{Var}(Y \mid S=s) = \mathrm{Var}(\langle \boldsymbol{X}, \boldsymbol{\beta^*} \rangle + \zeta \mid S=s)\enspace .
    $$
    By condition (iii), the distribution of $\boldsymbol{X}$ is independent of $S$, so this variance is a constant, $C$, for all $s$. Thus, 
    $$
    \mathbb{E}[\mathrm{Var}(Y\mid S)] = C\enspace.
    $$
    \item The conditional expectation is 
    $$
    \mathbb{E}[Y \mid S=s] = \mathbb{E}[\langle \boldsymbol{X}, \boldsymbol{\beta^*} \rangle + \beta_0^* \mid S=s]\enspace .
    $$
    By condition (iii), $\mathbb{E}[\boldsymbol{X} \mid S=s]$ is a constant vector, making $\mathbb{E}[Y \mid S=s]$ a constant value for all $s$.
\end{itemize}
It follows that $\mathrm{Var}(\mathbb{E}[Y\mid S]) = 0$. Therefore, 
$$
\mathrm{Var}(Y) = C + 0 = C\enspace,
$$
which means $\mathrm{Var}(Y) = \mathrm{Var}(Y \mid S=s)$ for all $s$.

With identical denominators, equation (*) simplifies to showing that:
\begin{align*}
    \mathrm{Var}&(Y - f(\boldsymbol{X})) \\ &= \sum_s p_{s} \mathrm{Var}(Y - f(\boldsymbol{X}) \mid S=s) \\ &=  \mathbb{E}[\mathrm{Var}(Y - f(\boldsymbol{X}) \mid S)]\enspace.
\end{align*}
Note we used condition (i) to write $f(\boldsymbol{X}, s) = f(\boldsymbol{X})$.

\medskip
\textit{Step B: Equality of numerators.} Let the residual be 
$$
R = Y - f(\boldsymbol{X})\enspace.
$$
The Law of Total Variance for $R$ is 
$$
\mathrm{Var}(R) = \mathbb{E}[\mathrm{Var}(R \mid S)] + \mathrm{Var}(\mathbb{E}[R \mid S])
\enspace.$$
The simplified equality from Step A holds if and only if $\mathrm{Var}(\mathbb{E}[R \mid S]) = 0$. This requires $\mathbb{E}[R \mid S=s]$ to be constant for all $s$.
\[
\mathbb{E}[R \mid S=s] = \mathbb{E}[Y \mid S=s] - \mathbb{E}[f(\boldsymbol{X}) \mid S=s]
\]
We already showed in Step A that $\mathbb{E}[Y \mid S=s]$ is constant. By condition (iii), the distribution of $\boldsymbol{X}$ is independent of $S$, so $\mathbb{E}[f(\boldsymbol{X}) \mid S=s]$ must also be constant for any function $f$. Since both terms are constant, their difference is constant, meaning $\mathrm{Var}(\mathbb{E}[R \mid S]) = 0$. The equality holds.

\paragraph{Part 2: ($\Rightarrow$)}
Assume equation (*) holds.
\begin{itemize}
    \item \textit{Condition (i) must hold.} If $f$ depended on $S$, the global error variance $\mathrm{Var}(Y - f(\boldsymbol{X},S))$ would contain a term related to the main effect of $S$ in the model $f$, whereas the group-level variances $\mathrm{Var}(Y - f(\boldsymbol{X},s) \mid S=s)$ would not. The functional forms would be different, making a robust equality impossible.
    \item \textit{Conditions (ii) and (iii) must hold.} If $\gamma^* \neq 0$ or if $\boldsymbol{X}$ is not independent of $S$, then as shown in Step A,  
    $$
    \mathrm{Var}(\mathbb{E}[Y\mid S]) > 0
    \enspace,$$
    and consequently 
    $$
    \mathrm{Var}(Y) > \mathbb{E}[\mathrm{Var}(Y\mid S)]
    \enspace.$$
    This creates a structural mismatch in the denominators of (*). Similarly, the conditional mean of the residual $\mathbb{E}[Y-f(\boldsymbol{X}) \mid S=s]$ would not be constant, creating a mismatch in the numerators where $\mathrm{Var}(Y-f) > \mathbb{E}[\mathrm{Var}(Y-f\mid S)]$. For the equality (*) to hold generally, these structural misalignments must be absent, which requires $\gamma^*=0$ and $\boldsymbol{X}$ to be independent of $S$.
\end{itemize}
This completes the proof.

\end{proof}

\subsection{Proof of Proposition \ref{prop: decomposition of the R2 Gap}} \label{sub: proof of decomposition of the R2 Gap}

\begin{proof}
We derive the expressions for $GWR^2$ and $R^2_{\mathrm{global}}$ separately under the stated assumptions.

First, we express the group-weighted R-squared. The R-squared for a specific group $s$ is:
\[
R^2_s = 1 - \frac{\mathrm{Var}(Y - f(\boldsymbol{X}) \mid S=s)}{\mathrm{Var}(Y \mid S=s)} = 1 - \frac{W_R}{W_Y}
\]
Since this value is constant for all groups, the weighted average is simply the value itself:
\begin{equation} \label{eq:R2_weighted_proof}
    GWR^2 = \sum_s p_{s} \left(1 - \frac{W_R}{W_Y}\right) = 1 - \frac{W_R}{W_Y}
\end{equation}

Next, we express the global R-squared. We apply the Law of Total Variance to the denominator, $\mathrm{Var}(Y)$, and the numerator, $\mathrm{Var}(Y - f(\boldsymbol{X}))$.
\begin{align*}
    \mathrm{Var}(Y) &= \mathbb{E}[\mathrm{Var}(Y \mid S)] + \mathrm{Var}(\mathbb{E}[Y \mid S])\\ &= W_Y + B_Y \\
    \mathrm{Var}(Y - f(\boldsymbol{X})) &= \mathbb{E}[\mathrm{Var}(Y - f(\boldsymbol{X}) \mid S)]\\ & + \mathrm{Var}(\mathbb{E}[Y - f(\boldsymbol{X}) \mid S]) \\ &= W_R + B_R
\end{align*}
Substituting these into the definition of global R-squared yields:
\begin{equation} \label{eq:R2_global_proof}
    R^2_{\mathrm{global}} = 1 - \frac{\mathrm{Var}(Y - f(\boldsymbol{X}))}{\mathrm{Var}(Y)} = 1 - \frac{W_R + B_R}{W_Y + B_Y}
\end{equation}

Finally, we compute the difference between \eqref{eq:R2_weighted_proof} and \eqref{eq:R2_global_proof}:
\begin{align*}
    GWR^2 - R^2_{\mathrm{global}} &= \left(1 - \frac{W_R}{W_Y}\right) - \left(1 - \frac{W_R + B_R}{W_Y + B_Y}\right) \\
    &= \frac{W_R + B_R}{W_Y + B_Y} - \frac{W_R}{W_Y} \\
    &= \frac{W_Y(W_R + B_R) - W_R(W_Y + B_Y)}{W_Y(W_Y + B_Y)} \\
    &= \frac{W_Y W_R + W_Y B_R - W_R W_Y - W_R B_Y}{W_Y(W_Y + B_Y)} \\
    &= \frac{W_Y B_R - W_R B_Y}{W_Y(W_Y + B_Y)}
\end{align*}
This completes the proof.
\end{proof}

\section{Numerical Experiments} \label{sub: appendix numerical experiments}

\subsection{Synthetic Dataset} \label{sub: synthetic data}

We provide additional details on the construction of the synthetic dataset used in our experiments. The data-generating process is governed by four control parameters $T :=(T_y,T_{\text{mean}},T_{\text{std}}, T_{\text{corr}})$. We generate synthetic triplets $(\boldsymbol{X},{S},Y)$ with the following properties:

\begin{itemize}
    \item \textbf{Sensitive attribute:} $S \in \{1,2\}$ follows a shifted Bernoulli distribution $\mathcal{B}(q)+1$ where $q=\mathbb{P}(Z > \tau)$ with $Z\sim \mathcal{N}(0,1)$ and fixed threshold $\tau$.
    \item \textbf{Features:} $\boldsymbol{X} \in \mathbb{R}^d$ is a multivariate Gaussian random vector whose distribution depends on the sensitive attribute $S$. The conditional distribution parameters are designed to introduce \textbf{indirect bias} through group differences in means, variances and correlation structures.
    \item \textbf{Outcome:} $Y=\sum_{j=1}^{d}X_j+T_y\cdot S$, where parameter $T_y$ directly controls the \textbf{direct bias} and introduces outcome disparity between groups.
\end{itemize}


\paragraph{Definition of the features X simulations:}
The conditional distribution of $\boldsymbol{X}$ given $S=s$ is $\mathcal{N}(\boldsymbol{\mu}^{(s)}, \boldsymbol{\Sigma}^{(s)})$, for each $s \in \mathcal{S}$, where the parameters are constructed as follows to systematically introduce different types of bias:

\begin{enumerate}
    \item \textbf{Mean vectors:} For each feature $X_j$, with $j \in \{1,\ldots,d\}$:
    \begin{align*}
        \mu_j^{(1)} &\sim \mathcal{B}(3,p) \quad \text{where } p \sim \mathcal{U}_{[0,1]} \\
        \mu_j^{(2)} &= \mu_j^{(1)} + T_{mean}
    \end{align*}
    The parameter $T_{mean}$ controls indirect mean bias by systematically shifting the mean of features for group 2 relative to group 1. When $T_{mean} = 0$, both groups have identical feature means, eliminating this source of bias.
    
    \item \textbf{Standard deviations:} For each feature $X_j$, with $j \in \{1,\ldots,d\}$:
    \begin{align*}
        \sigma_j^{(1)} &\sim \mathcal{U}_{[0,2]} \\
        \sigma_j^{(2)} &= \sigma_j^{(1)} + \sqrt{T_{std}}
    \end{align*}
    The parameter $T_{std}$ contributes to indirect structural bias by controlling differences in feature variances between groups. When $T_{std} = 0$, both groups have identical feature standard deviations.
    
    \item \textbf{Correlation matrices:} The correlation matrix $\boldsymbol{\rho}^{(s)}$ of $\boldsymbol{X}$ given $S=s$, for each $s \in \mathcal{S}$,  is constructed as follows  to control the correlation component of structural bias.
    
    For each group $s$, we generate a random matrix $\boldsymbol{A}^{(s)} \in \mathbb{R}^{d \times d}$ with i.i.d. entries $A^{(s)}_{ij} \sim \mathcal{N}(0,1)$.
    
    \begin{itemize}
        \item If $T_{corr} = 0$ (Independent features for both groups):
        $$\boldsymbol{\rho}^{(1)}=\boldsymbol{\rho}^{(2)}=\boldsymbol{I}_d\enspace.$$
        \item If $T_{corr} \in (0,1)$ (Group-Specific Correlations):
        \begin{align*}
            \boldsymbol{\rho}^{(s)} = T_{corr} \cdot \boldsymbol{C}^{(s)} + (1-T_{corr}) \cdot \boldsymbol{I}_d\enspace.
        \end{align*}
        where $\boldsymbol{C}^{(s)} = \boldsymbol{D}_s^{-\frac{1}{2}} (\boldsymbol{A}^{(s)\top} \boldsymbol{A}^{(s)}) \boldsymbol{D}_s^{-\frac{1}{2}}$ is the normalized correlation matrix derived from $\boldsymbol{A}^{(s)}$, $\boldsymbol{D}_s = \text{diag}(\boldsymbol{A}^{(s)\top} \boldsymbol{A}^{(s)})$ is the diagonal matrix of $\boldsymbol{A}^{(s)\top} \boldsymbol{A}^{(s)}$ and $\boldsymbol{I}_d$ is the identity matrix.
        \item If $T_{corr} = 1$ (Identical Correlations for both groups):
        \begin{align*}
            \boldsymbol{\rho}^{(1)} = \boldsymbol{\rho}^{(2)} = \boldsymbol{D}_1^{-\frac{1}{2}} (\boldsymbol{A}^{(1)\top} \boldsymbol{A}^{(1)}) \boldsymbol{D}_1^{-\frac{1}{2}}\enspace.
        \end{align*}
        Both groups share the same correlation structure.
    \end{itemize}
    
    The parameter $T_{corr} \in (0,1)$ defines the strength of correlation-based structural bias, while $T_{corr} \in \{0,1\}$ eliminates correlation-based structural bias.

    \item \textbf{Covariance matrices:} The final covariance matrix of $\boldsymbol{X}$ given $S=s$, for each $s \in \mathcal{S}$, is computed as:
    \begin{align*}
        \boldsymbol{\Sigma}^{(s)} = \text{diag}(\boldsymbol{\sigma}^{(s)}) \boldsymbol{\rho}^{(s)} \text{diag}(\boldsymbol{\sigma}^{(s)})\enspace,
    \end{align*}
    where $\text{diag}(\boldsymbol{\sigma}^{(s)})$ is the diagonal matrix containing the standard deviations $\sigma_j^{(s)}$.
\end{enumerate}

\paragraph{Systematic Bias Control.} This synthetic data generation process allows us to systematically control and isolate different sources of bias through the parameter vector $T$:
\begin{itemize}
    \item \textbf{Direct Bias} is controlled  by $T_y$ through explicit dependence of $Y$ on $S$.
    \item  \textbf{Indirect Mean Bias} is driven by $T_{mean}$ through systematic differences in $\boldsymbol{\mu}^{(s)}$.
    \item \textbf{Indirect Structural Bias} is controlled jointly by $T_{std}$ and $T_{corr}$ through group-specific differences in $\boldsymbol{\Sigma}^{(s)}$.
\end{itemize}

\subsection{Additional illustrations of coefficients adjustments for fairness in bias scenario}
Building on Section $7.1$, we further illustrate transparent bias remediation through coefficient adjustments. Using our synthetic dataset, we examine how our model adapts coefficients across three scenarios of increasing complexity, where each scenario adds new bias sources to the previous one: (1) Direct Bias, (2) Direct and Indirect Mean Bias, (3) Direct, Indirect Mean and Structural Bias.

\begin{figure}[htbp!]
    \centering
    \includegraphics[width=1\linewidth]{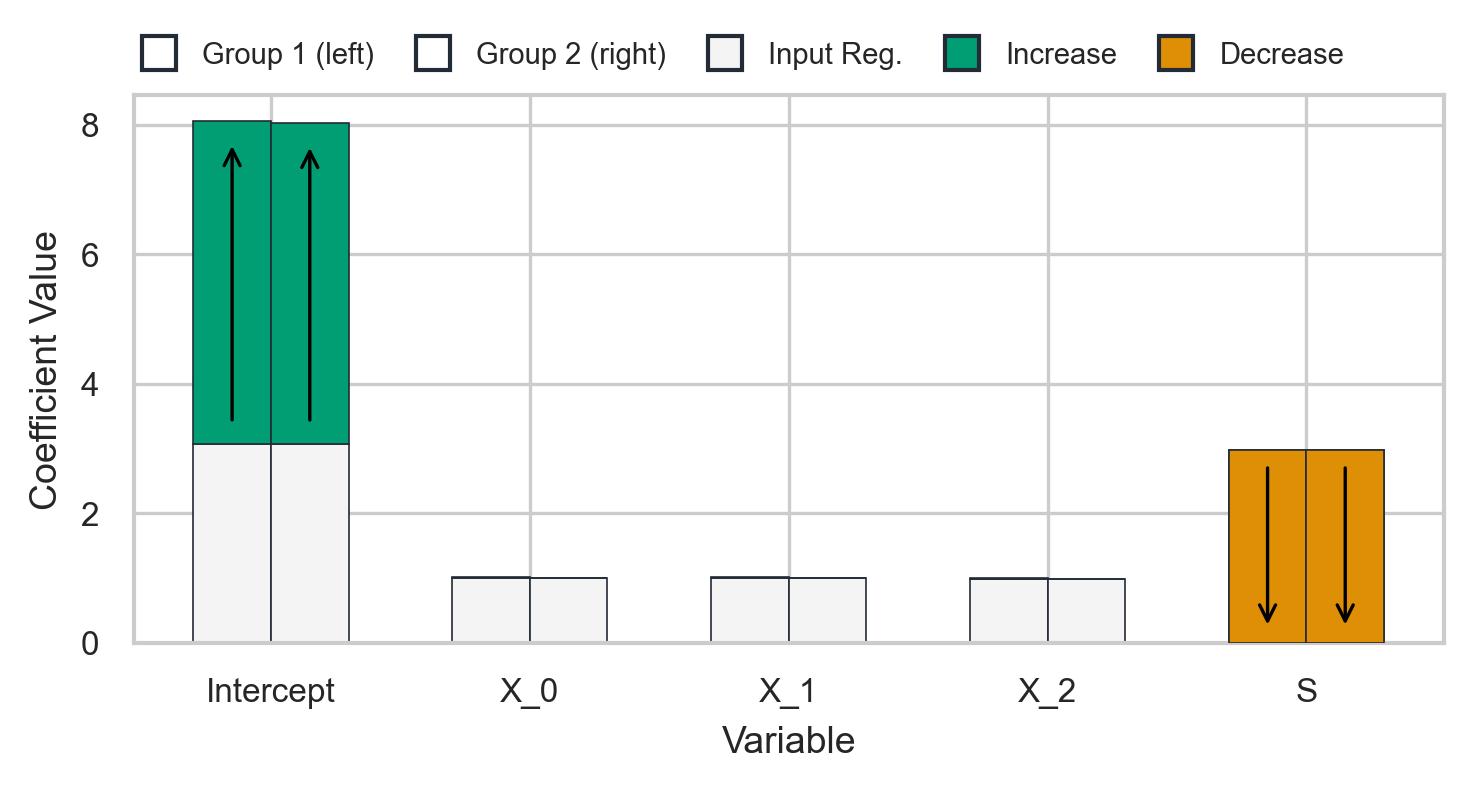}
    \caption{Coefficients adjustments for fairness in a direct bias scenario T $=(3,0,0,0)$, shown on synthetic data for a sample of features.}
    \label{fig: Coefficients adjustments in direct bias scenario.}
\end{figure}

\paragraph{Direct Bias.} In presence of direct bias only, our model eliminates unfairness through two adjustments: (1) it nullifies the sensitive attribute coefficient, removing direct dependence on group membership, and (2) it adjusts the intercept equally for both groups to maintain predictive accuracy (Fig. \ref{fig: Coefficients adjustments in direct bias scenario.}). 

\begin{figure}[htbp!]
    \centering
    \includegraphics[width=1\linewidth]{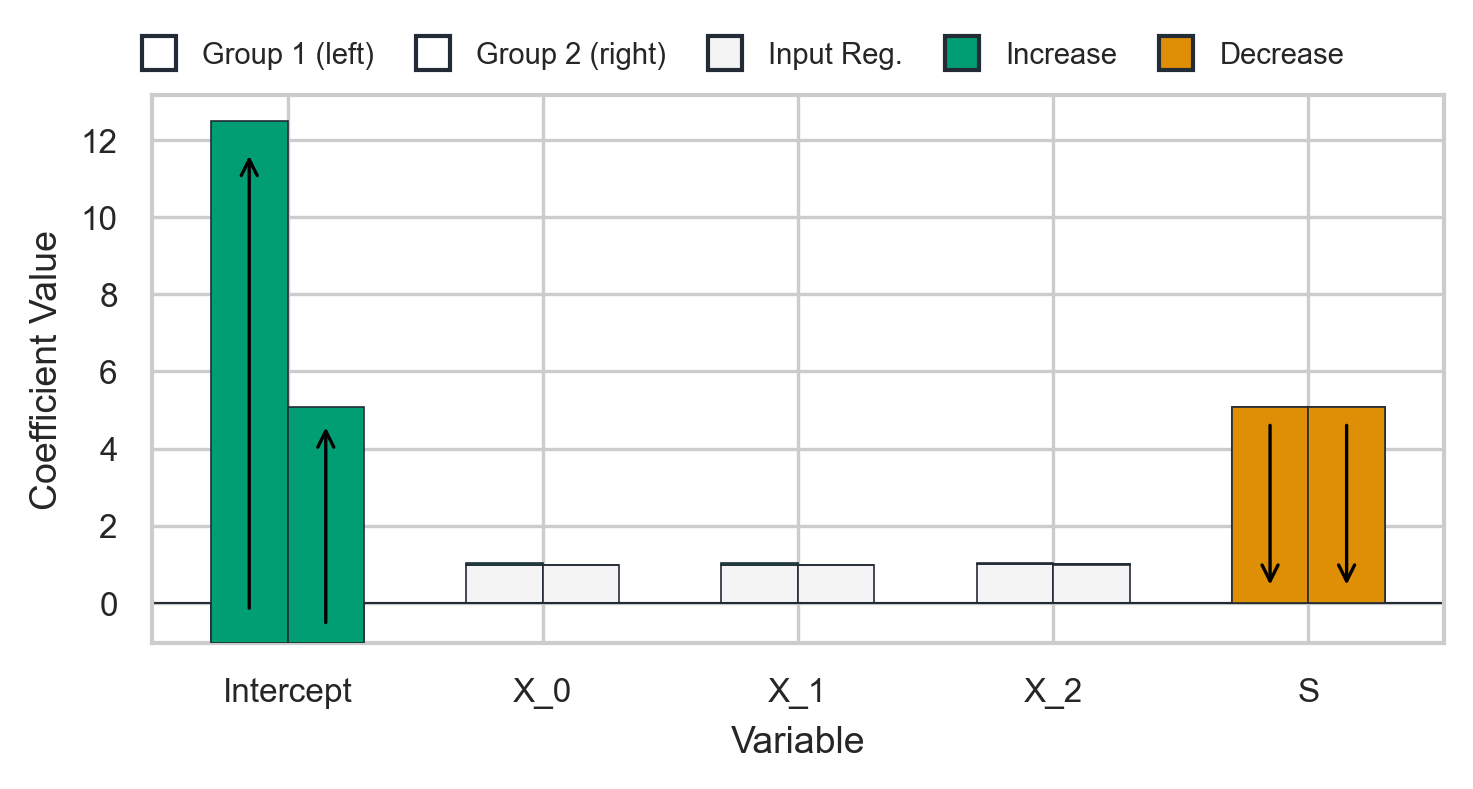}
    \caption{Coefficients adjustments for fairness in a direct and indirect mean bias scenario T $=(3,2,0,0)$, shown on synthetic data for a sample of features.}
    \label{fig: Coefficients adjustments in direct and indirect (mean) bias scenario}
\end{figure}

\paragraph{Direct and Indirect Mean Bias.} 
When indirect mean bias is added to direct bias, the remediation strategy becomes asymmetric. Beyond nullifying the sensitive attribute coefficient, the model compensates for the systematic difference in feature means between groups through unequal intercept adjustments. Group 1, which has lower feature means by construction ($\mu^{(1)} < \mu^{(2)}$ due to $T_{mean} > 0$), receives a larger positive intercept adjustment to offset this disadvantage and achieve fair outcomes (Fig. \ref{fig: Coefficients adjustments in direct and indirect (mean) bias scenario}).

\begin{figure}[b!]
    \centering
    \includegraphics[width=1\linewidth]{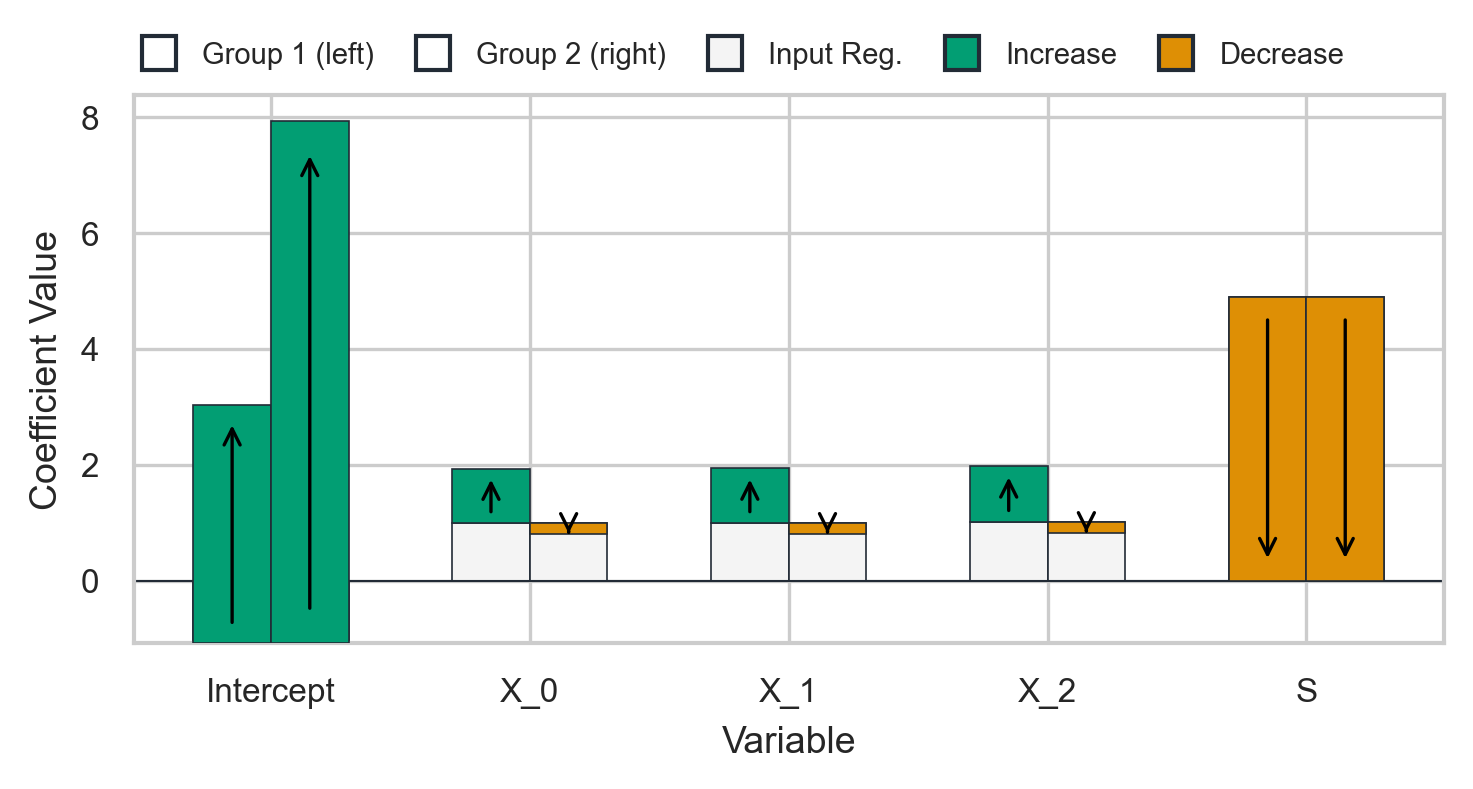}
    \caption{Coefficients adjustments for fairness in a direct and indirect (through mean and variance) bias scenario T $=(3,2,3,0)$, shown on synthetic data for a sample of features.}
    \label{fig: Coefficients adjustments in direct and indirect (mean and var) bias scenario}
\end{figure}

\paragraph{Direct, Indirect Mean and Structural Bias.}
Adding structural bias ($T_{std}>0$, $T_{corr}>0$) to the previously described biases activates the scaling factor $\frac{\bar{\sigma}_{f^*}}{\sigma^{(s)}_{f^*}}$ of our fair model. As a result, structural bias introduces two additional remediation mechanisms that build upon the existing direct and indirect mean bias corrections: (1) a new mechanism, the adjustment of the slope and (2) the further adjustment of the intercept. To analyze both effects clearly, we examine them in two stages: first considering variance differences alone, then incorporating both variance and correlation differences.
\begin{enumerate}
    \item \textbf{Adjustment of the slope:} The features' coefficients are now adjusted by the group-dependent scaling factor since $$\boldsymbol{\beta}^{(s)}_0=\frac{\bar{\sigma}_{f^*}}{\sigma^{(s)}_{f^*}}\boldsymbol{\beta}^*,$$ (as introduced in Perspective 2 with $\varepsilon=0$, within Section $4.4$). \begin{itemize}
        \item \textbf{Structural Bias through Variance:} Group 2's higher variance leads to coefficient slightly down-scaling, while Group 1's coefficients are up-scaled (Fig.~\ref{fig: Coefficients adjustments in direct and indirect (mean and var) bias scenario}). Due to the uniform variance structure (no correlations and constant variance differences across features), all coefficients within each group are adjusted by the same scaling factor, resulting in parallel shifts for all features.
        \item \textbf{Structural Bias through Variance and Correlations:} In a full bias scenario (Fig.~\ref{fig: Coefficients adjustments.}), all coefficients within each group are adjusted by a slightly different scaling factor due to the group correlations. 
    \end{itemize}
    \item \textbf{Modified adjustment of the intercept:} The intercept adjustment mechanism, observed in the previous scenario, is now modified because the scaling factor affects the mean-based correction. The intercept of our fair model (as introduced in Perspective 2 with $\varepsilon=0$, within Section 4.4) becomes :
    $$\beta_{0,0}^{(s)} = \bar{\mu}_{f^*}- \left(\frac{\bar{\sigma}_{f^*}}{\sigma_{f^*}^{(s)}}\right) \langle \boldsymbol{\mu}^{(s)}, \boldsymbol{\beta^*} \rangle \enspace.$$
    \begin{itemize}
        \item \textbf{Structural Bias through Variance:} In Fig.~\ref{fig: Coefficients adjustments in direct and indirect (mean and var) bias scenario}, due to Group 2 higher variance, the scaling factor $\frac{\bar{\sigma}_{f^*}}{\sigma^{(s)}_{f^*}}$ reduces the correction applied to its higher means.  Thus, Group 2 receives a slightly larger intercept's adjustment than Group 1, reversing the asymmetry observed in the previous scenario with direct and indirect mean bias (Fig. \ref{fig: Coefficients adjustments in direct and indirect (mean) bias scenario}). 
        \item \textbf{Structural Bias through Variance and Correlations:} Fig.~\ref{fig: Coefficients adjustments.} shows that group-specific correlation structures (randomly generated in our synthetic data) further modify the intercept adjustments, with Group 1 receiving additional corrections that reduce the asymmetry between both intercepts. 
    \end{itemize}
\end{enumerate}

Thus, unlike previous scenarios where adjustments were purely additive, structural bias remediation works through scaling effects.

This detailed coefficient analysis demonstrates the value of our framework's interpretability: practitioners can precisely diagnose bias sources, understand the specific remediation mechanisms at work, and make informed decisions about fairness interventions.

\end{document}